\documentclass[num-refs]{wiley-article}

\usepackage{siunitx}
\usepackage{booktabs}
\usepackage{graphicx}
\usepackage{algorithm}
\usepackage{algpseudocode}
\usepackage{amsmath}

\papertype{Original Article}
\paperfield{Journal Section}

\title{Vehicle Fuel Optimization Under Real-World Driving Conditions: An Explainable Artificial Intelligence Approach.}

\abbrevs{ABC, a black cat; DEF, doesn't ever fret; GHI, goes home immediately.}

\author[1, 2\authfn{1}]{Alberto Barbado}
\author[2]{Óscar Corcho}

\contrib[\authfn{1}]{Equally contributing authors.}

\affil[1]{Telefónica, 28050 Madrid, Spain}
\affil[2]{Departamento de Inteligencia Artificial, Universidad Politécnica de Madrid, 28660 Boadilla del Monte, Spain}

\corraddress{Telefónica, 28050 Madrid, Spain}
\corremail{alberto.barbadogonzalez@telefonica.com}

\presentadd[\authfn{1}]{Telefónica, 28050 Madrid, Spain}

\runningauthor{Barbado, A. and Corcho, O.}

\begin{document}

\begin{frontmatter}
\maketitle

\begin{abstract}
Fuel optimization of diesel and petrol vehicles within industrial fleets is critical for mitigating costs and reducing emissions. This objective is achievable by acting on fuel-related factors, such as the driving behaviour style. 

In this study, we developed an Explainable Boosting Machine (EBM) model to predict fuel consumption of different types of industrial vehicles, using real-world data collected from 2020 to 2021. This Machine Learning model also explains the relationship between the input factors and fuel consumption, quantifying the individual contribution of each one of them. The explanations provided by the model are compared with domain knowledge in order to see if they are aligned. The results show that the 70\% of the categories associated to the fuel-factors are similar to the previous literature.

With the EBM algorithm, we estimate that optimizing driving behaviour decreases fuel consumption between 12\% and 15\% in a large fleet (more than 1000 vehicles).

\end{abstract}
\end{frontmatter}

\section{Introduction}
Reducing fuel consumption within a fleet of vehicles from a company is critical, since it has impacts on several aspects, such as the economic costs, and for fuels such as petrol and diesel, it also has an impact on emissions. For example, for a company operating in Spain with 100 diesel vehicles that have an average fuel consumption per vehicle and month of 30 liters, the economic cost will be 3930 Euros, taxes included (considering the average price per liter of petrol in Spain: 0.609 Euros, without taxes; 1.31 Euros after taxes as of March 2021 \cite{GobEspFuelPrice}).

Together with that, it also has an environmental impact in terms of emissions (e.g. CO2) principally for diesel and petrol vehicles. Just in Spain, the average monthly fuel consumption for the automation sector (only diesel) is around 1.8M T (in December 2020) \cite{CNMCFuelSpain}. Considering an estimate of 2.67633 Kg of CO2 per liter of diesel \cite{DieselLitresTOCo2}, this translates into 4.82M T of C02 emissions each month. It is true that these emissions will be reduced by the transition to hybrid and electric vehicles. However, in US, by 2030, it is estimated that only the 7\% of the vehicles will be electric \cite{USATransitionElectric}. This highlights the need for finding complementary solutions in the meantime that help reducing vehicle emissions while they are progressively changed into electric ones.

The reduction of both economic costs and emissions is achievable by optimizing the fuel consumption of the individual vehicles of a fleet. This is something that, according to the literature, can be done by operating over different aspects that affect the fuel consumed by a vehicle during a route. For instance, \cite{zacharof2016review} indicates how the impact on fuel consumption by aggressive driving can be around the 26\% of the total fuel consumed by a vehicle. This means that simply optimizing the driving style of the drivers of a fleet has a significant direct impact on both the economic costs and emissions reduction.

The literature analysis on the variables that impact fuel consumption is useful in itself. However, it can be complemented using techniques that can automatically explain for an individual fleet what actionable features are impacting fuel consumption, and how much. This could be helpful for quantifying the potential economic saving and emissions reduction for that particular fleet. A set of techniques that answer this problem are Explainable Artificial Intelligence (XAI) algorithms.

Before XAI, there was a dichotomy on whether to use whitebox algorithms or blackbox ones to solve AI-related problems. Whitebox algorithms can directly explain the relationship between input and output features, in exchange of potential limitations on the modelling between those input and output features. For instance, a Linear Regression model is considered whitebox, but the modelling limitation is that the relationship inferred between input and output is linear. On the other hand, blackbox algorithms can potentially infer better relationships between input and output (e.g. by inferring non-linear relationships), but in exchange of not being able to provide clear explanations about those relationships.
XAI came to close this bridge by discovering ways to either apply algorithms that explain the relationships in a blackbox model, or by using new whitebox algorithms that can infer complex relationships between input and output. This last case is what happens with Explainable Boosting Machine (EBM). EBM is an algorithm that provides feature relevance based explanations (similar to a Linear Regression model) that allow to see the individual impact of the input features on the output.

XAI in general, and the EBM algorithm in particular, is seen as useful within the literature to understand the relationships between a set of input features and an output one. To the best of our knowledge, it has not been studied within the field of vehicle fuel consumption. Using an algorithm such as EBM can be useful for solving the problem aforementioned: understanding the impact that different actionable features have on the fuel consumption of a particular fleet. With that, fleet managers and fleet operators can discover potential ways for reducing economic costs while looking after the environment, contributing to environmental Sustainable Development Goals (SDG), like SDG11 and SDG12 \cite{euSDG2021}.

Following this, the main contributions of our work are:
\begin{itemize}
\item Use real-world industry data sets that represent different types of vehicles, gathered data from telematic devices connected to the vehicles for more than one year, in order to elicit up to 70 features that may have a potential impact on fuel consumption according to the State of the Art (SOTA). 
\item Design a solution that first trains the EBM algorithm over that input data, and then generates explanations that quantify the potential impact that the input factor may have on fuel consumption. These explanations are combined with business knowledge that aim to align them with prior domain expertise. This solution also includes how to evaluate the explanations from a domain knowledge point of view.
\item Measure both the predictive power of the EBM algorithm over this real-world input data related to fuel consumption, while also measuring the quality of the explanations in a quantitative manner using prior domain knowledge.
\item Quantify the potential impact that driving behaviour has in the different vehicle fleets considered through the explanations provided by the EBM algorithm.
\end{itemize}

The rest of the paper is organized as follows. First, we describe the related work in the area of factors that impact in fuel consumption, together with previous works regarding the usage of Machine Learning (ML) within the context of vehicle fuel consumption. Then, we describe our proposal, first explaining the EBM algorithm, and then explaining the system architecture that we will use for generating the explanations and quantify the impact of the different feature groups in the fuel consumption, together with how to evaluate the explanations from a domain knowledge point of view . Following this, we present an empirical evaluation using real-world industry IoT data belonging to different fleets of vehicles. We conclude, showing potential future research lines of work.

\section{Related Work}
\subsection{Factors for fuel consumption in a vehicle}
Fuel consumption can significantly vary from one vehicle to another, even when comparing two vehicles from the same make, model, year and fuel type. This is caused by different factors that may increase or decrease the amount of fuel consumed during the same trip. The literature contains many studies that identify these factors and assess how much fuel could be saved when they are optimized. This is something very relevant for fleet managers.

\begin{figure}[h!]
\centering
 \begin{tabular}{c@{\qquad}c@{\qquad}c}
\includegraphics[width=0.8\columnwidth]{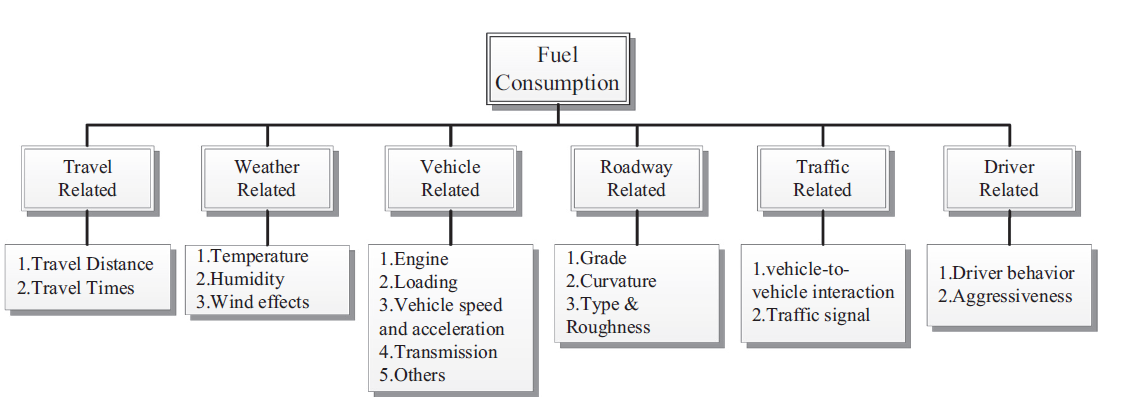}
  \end{tabular} 
  \caption{Categories of fuel factors discussed in \cite{zhou2016review} \label{figure:chart_factors}}
\end{figure}

\cite{zhou2016review} presents a literature review of different factors that have a potential impact in the fuel consumption of a vehicle, together with their relative importance. Figure \ref{figure:chart_factors} shows the categories of fuel factors considered in that review. 

The first category considered are \textbf{travel-related} factors. This group includes factors that are related to the route covered by the vehicle. In fact, the authors mention \textbf{eco-routing} as a crucial aspect to reduce fuel consumption. Fuel can be saved by choosing an optimal route not only in classical terms of distance and travel time, but also in terms of a route that saves fuel compared to other possible ones (e.g.choosing routes with less "bumps" or "slopes"). In fact, the new route may even be longer in time or distance, but offers fuel saving. The paper indicates that eco-routing alone can reduce the fuel consumption of a vehicle by 18\% to 23\%.

The second category includes \textbf{weather-related} factors. These factors impact the fuel consumption of a vehicle in an indirect way (i.e. by being related to the usage of the air conditioner, by affecting the water pump, by increasing the engine or transmission friction in a cold weather...). Thus, this category includes factors like the exterior temperature, the relative humidity or the wind effects. These factors may be responsible for about a 1\% of the fuel consumption of a vehicle.

The third group of factors are named \textbf{vehicle-related}. It includes factors mainly related to the engine and the vehicle itself, such as vehicle load, vehicle speed, engine speed, type of fuel, whether the vehicle has an exhaust after-treatment system or not... 

The fourth group is named \textbf{roadway-related factors}. It refers to factors related to the road condition, like the road slope, the surface roughness, or the road curvature. These factors, though not being very actionable (sometimes it is difficult to prevent them), have a large impact on the fuel consumption (around 5 to 20\%).

The fifth group of factors refer to \textbf{traffic conditions}. They are very related to a good arrangement of traffic signs, such as traffic lights. They have the potentially biggest fuel impact (around 22 to 50\% of the fuel consumption).

Finally, the sixth group mentioned in the review are the \textbf{driver-related} factors, like the driving behaviour or the aggressiveness of the driving. The driving profile of a particular driver (that measures aspects such as that driving aggressiveness), are calculated with vehicle information such as the RPM (engine speed; revolutions per minute), the speed or the acceleration. The authors mention how aggressive driving can be responsible for up to 40\% of the fuel consumption of a vehicle when compared to a calmer driving style. 

The aforementioned literature review is enhanced by the study of \cite{zacharof2016review}. Here the authors present a thorough analysis regarding the influence of different factors for fuel consumption in a vehicle, along with the influence for CO2 emissions. This study considers passenger vehicles under real-world operating conditions. Regarding fuel consumption specifically, the authors offer a summarized view of the literature showing different categories of variables and their proportional impact in the fuel consumption of a vehicle. 

There are two approaches for analysing the impact of a specific factor in the fuel consumption of a vehicle. First, using a simulation analysis that studies the isolated impact of a factor under laboratory conditions. Second, by analysing feeds of data that contain the instant fuel consumption reported during trips on real-world environments. These feeds of data can be gathered from sources such as OBD-II (On-board diagnostics) port \cite{ISO14230} (e.g. the Engine Fuel Rate with the PID 015E). 

The analysis of the literature highlights that both approaches offer in general similar results (when there are publications available for a specific factor both from the simulation point of view, as well as with the real-world data). Thus, real-world collected data can be a valid data source for assessing the impact of different factors in the fuel consumption of a vehicle.

Here, the literature review proposes a fuel factor taxonomy that in some cases matches directly the one proposed in \cite{zhou2016review}, but in some others is different. There are 28 factors than can be classified into 9 groups. All these factors, as reported by \cite{zacharof2016review}, appear at Table \ref{table:factors_table}. This Table shows the relative importance of each of the factors (literature median value) along with an interval that encloses the different values reported, considering vehicles under real-world operating conditions. It also shows how many papers talk about that particular factor, as well as the distribution of the relative values reported. 

\begin{table}[]
\centering
 \begin{tabular}{c@{\qquad}c@{\qquad}c}
\includegraphics[width=0.95\columnwidth]{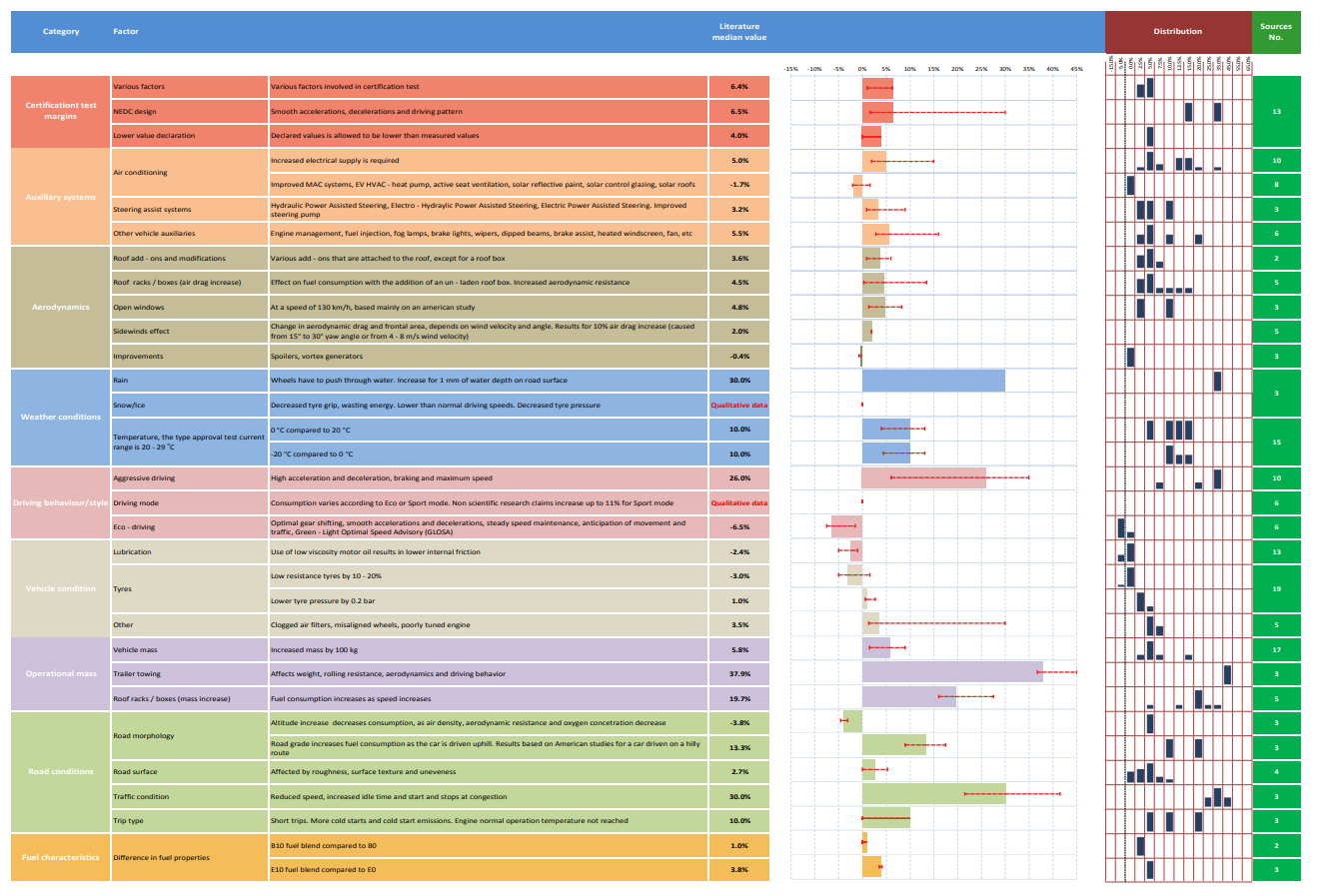}
  \end{tabular} 
  \caption{Fuel factors mentioned in the literature, together with the relative importance as reported by  \cite{zacharof2016review} \label{table:factors_table}}
\end{table}

Regarding driver-related factors, Table \ref{table:factors_table} shows a group called \textbf{driving behaviour/style} that accounts for factors related directly to the driver. It is almost similar to the one from \ref{figure:chart_factors} with the exception of considering factors related to good driving styles that may reduce the fuel consumption.

Regarding the group \textbf{road conditions} in \cite{zhou2016review}, it mainly includes the travel related, traffic related and roadway related factors.

Vehicle-related is the group with more factor's differences between both papers. Compared to \cite{zhou2016review}, these factors are split into \textbf{auxiliary systems}, \textbf{vehicle conditions} and \textbf{fuel characteristics}, complemented with other groups that include factors related to the vehicle's design itself (\textbf{aerodynamics} and \textbf{operational mass}) and to \textbf{certification test margins}. In this last review, all these vehicle-related factors account for aspects related to the vehicle itself, not considering anything directly related to the driver. This is a difference when compared to the taxonomy of \cite{zhou2016review}, because vehicle-related includes acceleration and speed factors.

The difference between the analyses shown in both articles are not only in terms of the taxonomy proposed to group factors, but sometimes also regarding the reported impact (i.e. exterior temperature has a median impact reported value of 10\% at \cite{zacharof2016review} against the 1\% impact for all weather related causes reported by \cite{zhou2016review}).

Within this last taxonomy of features that affect the fuel usage of a vehicle, some of them could be considered as "actionable", thus, they could be changed in a particular vehicle; in some cases without even needing to change the vehicle's route. An example of this is the aggressive driving style. Other features are inherent to the vehicle and cannot be directly changed, like the vehicle make/model or the vehicle mass. Even within the "actionable" features, some of them cannot be easily read through OBD-II (e.g., if there are roof add-on, which affects the vehicle aerodynamics). Thus, a subset of these features that considers only the ones that are "actionable" and the ones that can be read is the one shown in Figure \ref{table:FeatureInfluenceReduced}.

\begin{table}[]
\centering
 \begin{tabular}{c@{\qquad}c@{\qquad}c}
\includegraphics[width=0.90\columnwidth]{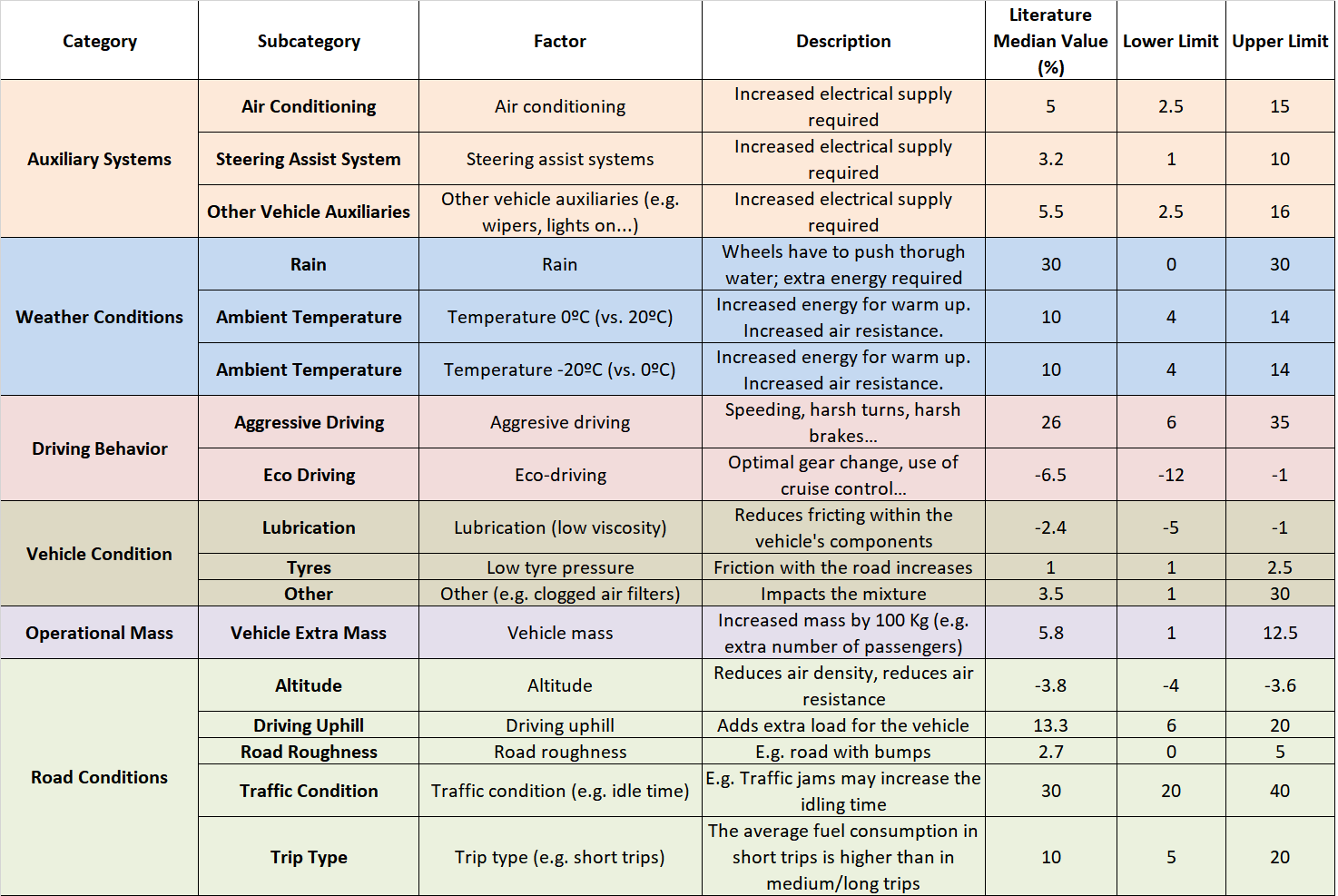}
  \end{tabular} 
  \caption{Reduced view from the factors of \cite{zacharof2016review}, focusing on some of the actionable variables that can be retrieved from the OBD-II. The upper and lower limits refers to the minimum and maximum SOTA values reported in the review. For Rain, the lower limit is set to zero since the review does not provide limits for that feature. \label{table:FeatureInfluenceReduced}}
\end{table}

The physical reasons as to why these features impact the fuel usage are:
\begin{itemize}
\item \textbf{Air conditioning (A/C)}: Using A/C increases the energy supply needed, leading to an increased fuel consumption. The time using the A/C and the power needed will increase/decrease that extra energy required. This category also includes the heating systems and related features, like the vehicle's coolant.
\item \textbf{Steering assist system}: These systems help driving safely and more confortable, but require additional electrical supply in exchange. An example is the usage of Electric power assisted steering (EPAS).
\item \textbf{Other vehicle auxiliaries}: These features include other auxiliary elements of the vehicle that may also require an extra energy. An example is the vehicle lights usage, that require extra energy and due to that, extra fuel.
\item \textbf{Rain}: Rain (and snow) impact the fuel usage in different ways. First, they affect the wheel gripping to the road surface. Also, the wheels have to push through an additional layer of water (or snow), so extra energy is required.
\item \textbf{Ambient temperature}: Temperature affects tyres, motor oil viscosity, cold start engine… Extra fuel is required in low temperatures to warm up the engine. It also affects aerodynamics: increased air density and higher aerodynamics resistances.
\item \textbf{Aggressive driving}: Aggressive driving is shown through different variables: acceleration patterns, gear change, harsh turns, harsh brakes, speeding... The impact on the fuel usage could be high.
\item \textbf{Eco driving}: Eco driving is related to the optimal driving of a vehicle, which may reduce its fuel usage. It involves optimizing the gear shifting (related to the usage of cruise control), choosing the best possible route thanks to a navigation device...
\item \textbf{Lubrication}: Overcoming of friction within the vehicle's components requires energy, and this is related to the fuel usage. If the friction is minimized thanks to an adequate lubrication, the energy required will be lower.
\item \textbf{Tyres}: Tyre pressure is related to the rolling resistance coefficient (RRC). When the tyres have low pressure, the contact surface with the road increases and more energy is needed to rotate the wheel (as the friction increases). 
\item \textbf{Other (vehicle condition)}: Beside tyres and lubricants, there are other vehicle conditions that impact the fuel usage. For instance, if the air filters are clogged. This is something that happens mainly in old models (since fuel injection in new cars is adjusted to ensure the correct mixture). Other examples are misaligned wheels and suspension losses. 
\item \textbf{Vehicle extra mass}: Extra mass in a vehicle (measured, for instance, in additional 100Kg), increase the energy needed to move the vehicle. This may happen for instance when there are additional passengers in a vehicle.
\item \textbf{Altitude}: In higher altitudes the air density is lower, so the air resistance that the vehicle faces while driving is also lower. This means that in higher altitudes the vehicle needs lower energy to move the same distance.
\item \textbf{Driving uphill}: Driving uphill adds an extra load over the vehicle, that needs additional energy to move. By contrast, driving downhill reduces the amount of energy needed.
\item \textbf{Road roughness}: For instance, if a road has many bumps, the vehicle will need additional energy to go through it. 
\item \textbf{Traffic condition}: Traffic condition also impacts in the fuel usage. For instance, if there are traffic jams, the idling time normally increases, leading to an increased average fuel consumption.
\item \textbf{Trip type}: The trip type also impacts in the fuel usage. For instance, if the trip distance is small, the average fuel consumption will increase, since fuel is required to turn on the vehicle.
\end{itemize}

There are some additional factors that impact in the fuel consumption that the previous references did not mention. This is the case of Diesel Exhaust Fluid (DEF). DEF is an urea-based product used in after-treatment processes of the vehicle, such as Selective Catalytic Reduction (SCR). It is applied over the vehicle's exhaust stream in order to transform the NOx gas emissions into nitrogen, water and CO2, reducing the NOx emissions in the process \cite{betageri2016effects}. Techniques like SCR do not only reduce the emissions of a vehicle, but also help the engine performance and may lower fuel consumption \cite{chen2015nonlinear, chen2013integrated}.

The factors already mentioned are linked to passenger vehicles, but for other vehicles, such as trucks, there are additional ones to consider. This is the case of power take-off, where there is power from the engine that is taken out (e.g. with a splined drive shaft) and used in another application (e.g. for a cement mixer in a truck). This directly impacts in the mileage of a vehicle \cite{boriboonsomsin2010analysis}.

All these references show that there is a physical and empirically measured connection between the value of specific factors and the value of the fuel consumption. Thus, it is possible to use them in order to predict the value of the fuel consumption with ML models, as already shown within the literature \cite{9072728, 8727915, perrotta2017application}.

\subsection{Machine Learning for connecting input features to vehicle fuel consumption}
As we mentioned in the previous subsection, there are several features that affect the fuel consumption of a vehicle. This can be measured using as input data source the feeds of data gathered from the vehicle's movement together with Machine Learning (ML) algorithms. This is the case of \cite{ping2019impact}, where the authors conduct an study over a fleet of vehicles where they assess the impact of driving behaviour in the fuel consumption. They consider features related to driving behaviour, such as the gas pedal position, the speed and speed variance, or the steering angle, and they first see how those features have significant correlations with the fuel consumption. Then, they use several clustering algorithms (Spectral clustering, KFCM, K-Means), finding different clusters based on the driver consumption profile and its relationship with those driving behaviour features. 

In \cite{perrotta2017application}, the authors analyse the impact of other features for fuel consumption within the context of trucks. The 56 features used include characteristics from the vehicle, such as its gross weight, together with others belonging to driving behaviour (usage of cruise control, average speed...), as well as information from the road (like the road surface macrotexture, or the curvature of the road). Those input features are seen as correlated with the fuel consumption (using a bivariate correlation analysis), and then are used to train several ML models (ANN, SVM, Random Forest) in order to predict the fuel consumption of the trucks. For the case of Random Forest, the authors viewed the relative impact from the different features in the fuel consumption through their contribution for accuracy during the tree splitting process.

The previous approaches are useful for detecting dependencies between a set of features and the fuel consumption of a vehicle. However, they do not quantify exactly how many extra liters of fuel are spent due to those features. In \cite{andrieu2014evaluation}, the authors investigate the impact of eco-driving in the fuel consumption. Eco-driving is expressed through several features related to variables such as the Revolutions Per Minute (RPM) or the braking. Then, they use statistical tests for detecting significant decreases in fuel consumption when an eco-routing driving style is used. Then, they use a Logistic Regression model for analysing the relationship between driver-related features and the fact that the vehicle trip was actually done with eco-routing.

It is possible to use a Linear Regression model for measuring the individual impact of input features in fuel consumption, and know exactly how many liters are used due to each individual variable. The reason behind this is that those models are known as whitebox because they directly provide the influence of the input in the output \cite{alej2019explainable}.
This is shown in \cite{pavlovic2020understanding}, where the authors predict the fuel consumption gap between type-approval tests and real-world driving trips, using the information of one vehicle during one year, and with 20 different drivers. With that, they build a multiple linear regression model that takes into account driver-related factors as well as environmental and traffic factors in order to predict the fuel consumption gap. Through these linear models, they provide the relative importance for each of the features in the fuel consumption, as well as the r2 value for each of the models tested in order to evaluate them. Similarly, in  \cite{lasocki2019environmental} the authors study the impact on the fuel of several features inferred related to driving behaviour through the analysis of the data from two different vehicles. One of these features is the Driving Style Indicator (DSI), which is the difference between the average positive acceleration of a vehicle minus the average of the negative acceleration divided by the average speed. The relationship between these features and fuel consumption is modeled through linear regression algorithms in order to quantify the impact of each one of them.

Even though linear regression models can be used for fuel prediction when there is a need of a whitebox ML algorithm that explains the relationship between input and output, this limits the results since the relationship inferred is linear.

This problem can be solved by using non-linear whitebox models, such as Generative Additive Models (GAM). These models, instead of modelling the relationships between the input features and the output value though a constant coefficient, they infer individual and additive relationships through non-linear functions. 
This has been proved useful in other domains. In \cite{zhou2020identifying}, authors identify risk factors and interaction effects in order to to predict intensive care admission in patients hospitalized with COVID-19. In \cite{decroos2019interpretable}, authors use GAM models to predict goals in soccer along with the quantification of the impact of the different input factors on the output. However, to the best of our knowledge, these models have not not been used for both predicting fuel consumption along with the quantification of factors that impact on it.

\section{Method}
In this Section we describe the XAI whitebox algorithm considered in this work for obtaining the explanations, as well as the logic used for generating them. We also include the schema for the whole process, and the steps involved for analysing and evaluating those explanations.

\subsection{Explainable Boosting Machine (EBM)}
EBM \cite{nori2019interpretml} is a type of whitebox model that provides feature-relevance based explanations. It can be used for both regression and classification tasks, and similarly to other whitebox models, such as Linear or Logistic regression algorithms, it infers an independent relationship between input features and the output variable. Because of that, it is possible to know the individual contribution from those input features for a particular output value.
The advantage of EBM is that it provides the option to infer non-linear relationships, and due to that, it can potentially increase the model generalization \cite{molnar2019interpretable}. EBM is based on the $GA^2M$ algorithm \cite{lou2013accurate}, but with a difference in terms of computation performance. EBM is an evolution from Generalized Additive Model algorithms (GAM) \cite{hastie1987generalized} because not only it is able to model individual relationships between the input features and the output, but it can also model pairwise interactions between input features, and include them as additional terms. The expression for the EBM algorithm appears in Equation \ref{eq:EBM-pairwise} for the regression case. In that Equation, $\sum_{n=1} f_{i}(x_{i})$ represents the different functions that model the individual relationship between a specific input feature $x_{i}$ and the output $y$ through a specific link function $g$. Similarly, $\sum_{n=1} f_{ij}(x_{i}, x_{j})$ represents the pairwise function term that models the relationship between two input features $x_{i}, x_{j}$ and the output $g(E[y])$. Finally, $\beta_0$ represents the intercept that adjusts the prediction from the model. For the sake of simplicity in both the training and the explanations generated, we have not considered pairwise interaction terms for the analysis carried out in this paper (the hypeparameters of the model allows to choose whether or not to include them).

\begin{equation}\label{eq:EBM-pairwise}
g(E[y]) = \beta_0 + \sum f_{i}(x_{i}) + \sum f_{ij}(x_{i}, x_{j})
\end{equation}

\subsection{Explanation generation}
\begin{figure}[]
\centering
 \begin{tabular}{c@{\qquad}c@{\qquad}c}
\includegraphics[width=0.65 \columnwidth]{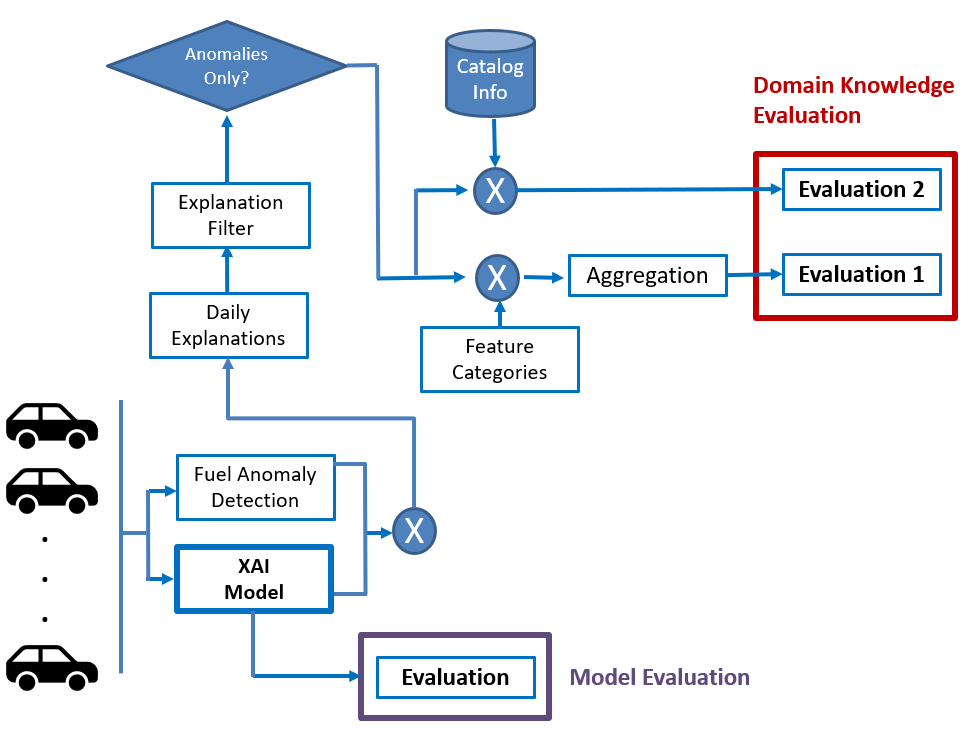}
  \end{tabular} 
  \caption{Schema for the process of aggregating the explanations in order to compare them with the SOTA.  \label{figure:SchemaProcessFuelCompare}}
\end{figure}

In this Subsection we show the full process followed for obtaining the explanations and for evaluating them later. This schema appears in Figure \ref{figure:SchemaProcessFuelCompare}. 

Using the databases with vehicle information (described in more detail in Subsection 4.1), we both identify fuel outliers and we also train the XAI whitebox model that infers the relationship between input features and fuel consumption. 

Regarding the anomaly detection step, it detects vehicles and dates where the average fuel consumption is anomalous, considering the other vehicles from the same model and for the dates that are associated to the same route type. That route type is the primary route type of a vehicle in a day (highway, city or combined). The algorithm applied is described in Equation \ref{eq:anomalies-fuel}. Using the whiskers from a boxplot analysis, the vehicle-dates with anomalous fuel consumption are detected with an univariate approach considering the average fuel consumption from other vehicles from the same model and route type.

\begin{equation}\label{eq:anomalies-fuel}
\begin{split}
lim\_sup = Q3 + 1.5 \times IQR \\
lim\_inf = Q1 - 1.5 \times IQR
\end{split}
\end{equation} 

\begin{figure}[]
\centering
 \begin{tabular}{c@{\qquad}c@{\qquad}c}
\includegraphics[width=0.65 \columnwidth]{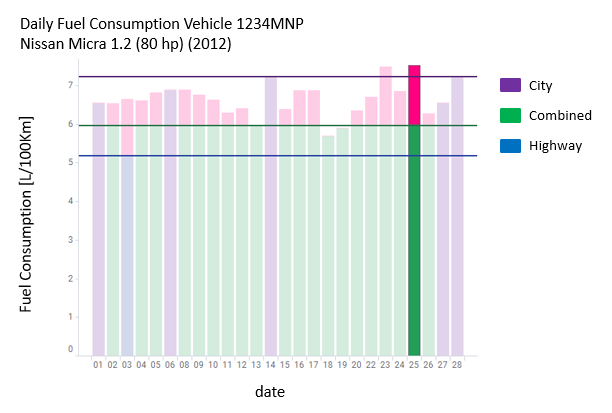}
  \end{tabular} 
  \caption{Output example for the fuel anomaly detection for one vehicle. It shows the maximum limit for that vehicle model with respect to the three route types (city, combined, highway), highlighting the fuel part that is anomalous because is above those limits.  \label{figure:AnomalyDetectionExample}}
\end{figure}

This approach is useful since it directly provides a fuel threshold that indicates the amount of fuel that is anomalous for a particular vehicle. In this paper we will only use the upper limit for the context of the explanations, since it is the one that identifies high fuel consumption. the An example of these limits appear in Figure \ref{figure:AnomalyDetectionExample} for one vehicle. We see that the different dates are categorized as "city", "combined", and "highway", and there is a threshold that highlights the amount of fuel that in some of those days is anomalous for that vehicle.

The information provided by the anomaly detection step will be used later on the evaluation step to analyse how much of the anomalous extra fuel is covered by the XAI explanations.

\begin{figure}[]
\centering
 \begin{tabular}{c@{\qquad}c@{\qquad}c}
\includegraphics[width=0.85 \columnwidth]{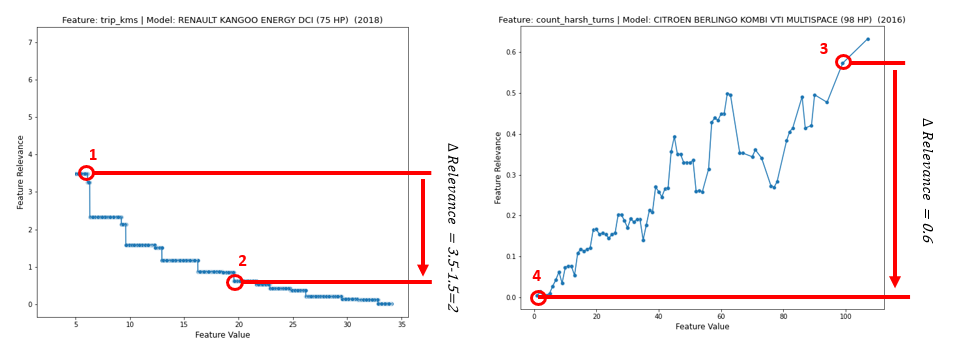}
  \end{tabular} 
  \caption{Example of the explanation generation. The first image shows the feature relevance for trip kms, and the second one for harsh turns. The explanations for points 1 and 3 corresponds to the feature relevance difference with respect to the reference points 2 and 4 respectively.  \label{figure:ExpMethodExample}}
\end{figure}

Regarding the XAI model, after training it,  we get its raw explanations (Daily Explanations step). There are several combinations of vehicle-date-feature where we have the feature relevance for each feature, for each vehicle and for each date. However, since the functions that express the relationship between input features and the output are not linear, we calculates the average fuel reduction for each vehicle and each date if every one of those features changed from its actual value to a reference value (e.g. if the tyre pressure increases from its actual value to the median value that it usually has for the vehicles of the same model). An example of this appears in Figure \ref{figure:ExpMethodExample}.

Thus, for every vehicle-date-feature we see how the feature relevance changes when the feature value changes from the current one to another reference value. This reference value can be either zero (for features identified with "Yes" in the "Reference Zero" column in Tables \ref{table:FARUsedpt1} and \ref{table:FARUsedpt2}) or the median feature value for the vehicles from the same model that have a fuel consumption that is not anomalous (according to Equation \ref{eq:anomalies-fuel}).
Equation \ref{eq:y-red-fuel} shows that fuel reduction for a particular vehicle in a specific date. It adds the fuel reduction from every feature $i$ within the explanation (from a total of N features). $x(i)$ is the feature value for that feature, $f_{i}(x(i))$ its feature relevance, $r$ is the route associated for that date, and $m$ being the vehicle's corresponding model. The Equation also shows how the reference value $x_{ref}$ changes.

\begin{equation}\label{eq:y-red-fuel}
\begin{aligned}
y\_fuel\_saved = \sum_{i=0}^{N} {f_{i}(x(i)) - x_{ref}}
\\
x_{ref} =
    \begin{cases}
      0 & \text{if $i$ in \textit{Reference Zero}}\\
      Median(X[r,m,i]) & \text{otherwise}
    \end{cases}     
\end{aligned}
\end{equation}

An example of the explanation output appears in Table \ref{table:data-explanation-sample}, where there is a row for each vehicle (vehicle\_id), date (date\_tx) and feature (e.g. the number of jackrabbit events, count\_jackrabbits). The variable vehicle\_group indicates the id associated to the group (make, model, year and fuel type) for that vehicle\_id. route\_type indicates the type of route for that specific date ("city", "combined" or "highway"). Complementing this, "avg\_fuel\_consumption" indicates the average fuel consumption for that vehicle in that date (L/100Km), and limit\_group the threshold limit that identifies when the average fuel consumption for a particular vehicle is anomalous (L/100Km). Along with that, it includes the intercept. Then, feature\_relevance contains the feature relevance for that vehicle-date, and feature\_value its corresponding value. target\_value shows the recommended value for changing that particular feature, and following that, y\_diff shows the average fuel consumption (L/100Km) that would be reduced by doing that. y\_fuel\_new shows the new average fuel consumption (L/100Km) that would be achieved by applying all the recommendations from the explanations in a particular day.

\begin{table}[]
\centering
\resizebox{\textwidth}{!}{%
\begin{tabular}{@{}llllllllllllll@{}}
\toprule
\textbf{vehicle\_id} &
  \textbf{date\_tx} &
  \textbf{route\_type} &
  \textbf{vehicle\_group} &
  \textbf{intercept} &
  \textbf{feature} &
  \textbf{feature\_relevance} &
  \textbf{feature\_value} &
  \textbf{target\_value} &
  \textbf{avg\_fuel\_consumption} &
  \textbf{limit\_group} &
  \textbf{y\_pred} &
  \textbf{y\_diff} &
  \textbf{y\_fuel\_new} \\ \midrule
id1 & 17/04/2020 & highway  & 0  & 7.25 & mean\_forward\_acc        & 0.2087 & 2.41   & 0.48  & 9.96 & 9.35 & 10.39 & 0.22 & 8.31 \\
id1 & 17/04/2020 & highway  & 0  & 7.25 & count\_jackrabbit         & 0.0568 & 9.0    & 0.0   & 9.96 & 9.35 & 10.39 & 0.06 & 8.31 \\
id1 & 17/04/2020 & highway  & 0  & 7.25 & mean\_speed\_hwy          & 1.0465 & 99.5   & 75.73 & 9.96 & 9.35 & 10.39 & 1.18 & 8.31 \\
id1 & 17/04/2020 & highway  & 0  & 7.25 & mean\_exterior\_temp      & 0.2581 & 282.65 & 287.9 & 9.96 & 9.35 & 10.39 & 0.07 & 8.31 \\
id1 & 17/04/2020 & highway  & 0  & 7.25 & count\_harsh\_turns       & 0.0947 & 11.0   & 0.0   & 9.96 & 9.35 & 10.39 & 0.12 & 8.31 \\
id56 & 30/04/2020 & combined & 14 & 7.25 & count\_neutral            & 0.0048 & 8.0    & 0.0   & 8.3  & 7.89 & 9.07  & 0.09 & 6.49 \\
id56 & 30/04/2020 & combined & 14 & 7.25 & count\_jackrabbit         & 0.0195 & 3.0    & 0.0   & 8.3  & 7.89 & 9.07  & 0.03 & 6.49 \\
id56 & 30/04/2020 & combined & 14 & 7.25 & rpm\_red                  & 0.0512 & 10.0   & 0.0   & 8.3  & 7.89 & 9.07  & 0.12 & 6.49 \\
id56 & 30/04/2020 & combined & 14 & 7.25 & rpm\_yellow               & 0.1954 & 17.0   & 0.0   & 8.3  & 7.89 & 9.07  & 0.22 & 6.49 \\
id56 & 30/04/2020 & combined & 14 & 7.25 & rpm\_orange               & 0.2774 & 30.0   & 0.0   & 8.3  & 7.89 & 9.07  & 0.36 & 6.49 \\
id56 & 30/04/2020 & combined & 14 & 7.25 & count\_speed\_limit\_90   & 0.154  & 154.0  & 17.0  & 8.3  & 7.89 & 9.07  & 0.15 & 6.49 \\
id56 & 30/04/2020 & combined & 14 & 7.25 & rpm\_high                 & 0.4663 & 179.0  & 0.0   & 8.3  & 7.89 & 9.07  & 1.48 & 6.49 \\
id56 & 30/04/2020 & combined & 14 & 7.25 & mean\_side\_to\_side\_acc & 0.0402 & 1.05   & 0.83  & 8.3  & 7.89 & 9.07  & 0.04 & 6.49 \\ \bottomrule
\end{tabular}%
}
\caption{Example of the explanations provided by the system.}
\label{table:data-explanation-sample}
\end{table}

A final note is that since the explanations show potential fuel savings when we change a feature value to a target one, the feature reduction will be obtained in the same one regardless of whether we need to increase the feature value or decrease it (ash shown in Figure \ref{figure:ExpMethodExample}).

After that previous step, we apply several business rules for filtering some of the explanations generated (Explanation Filter step). These rules are:
\begin{itemize}
\item BR1: The features used for training the model may be numeric (e.g. time driving uphill) or categorical (e.g. the vehicle model). All those categorical features are one-hot encoded before training the model. However, they are not considered for the explanations since they are not actionable.
\item BR2: We remove the features in the vehicle-date explanations that have a very low impact on the fuel consumption (relative impact below 1\%)
\item BR3: The explanations only include vehicles where the average fuel consumption is above the value of the median inlier vehicles for the same model and on the same route type.
\item BR4: Feature values must be higher than the median value of the vehicle inliers from the same model for that same feature when the feature Type is Positive, or lower when Type is Negative.
\item BR5: The total fuel reduction from the explanations should not be more than the 80\% of the original fuel consumption. Since EBM does not allow to impose restrictions in the learning for the individual models for the features, we need to apply this posthoc filtering to remove explanations that are not physically possible.
\end{itemize}

Finally, the schema shown in Figure \ref{figure:SchemaProcessFuelCompare} includes several evaluation steps. First, a "Model Evaluation", where the performance of the model is measures just like any other ML regression algorithm. We analyse the performance with the following metrics: R2 and the Mean Average Precision Error (MAPE). 

Along with this, we propose novel evaluation steps that evaluates the explanations against prior domain knowledge. The quantitative evaluation of XAI explanations in order to see if they are aligned with prior domain knowledge is what some authors have identified as the property of being "consistent with a priori beliefs" \cite{carvalho2019machine, barbado2019rule}. 

Following this, first, we use the individual explanations generated by the aforementioned solution, and after applying BR1, BR2 and BR3, we aggregate the relative impact in the average fuel consumption following the categories described in Table \ref{table:FeatureInfluenceReduced} from \cite{zacharof2016review} in order to see if the impact per category is aligned with the SOTA (Evaluation 2 in Figure \ref{figure:SchemaProcessFuelCompare}). Second, we compare the new average fuel consumption (after applying the recommendations from the Daily Explanations step) to the catalog fuel reference for vehicles of the same make, model, year, fuel type and on that specific route type (city, combined or highway). The intuition behind this is that if we remove the extra fuel due to driving behaviour, traffic conditions... the vehicle's fuel should be close to its catalog reference. The databases that we use to get this catalog fuel reference are  \cite{CAFuel2021}, \cite{USAFuel2021}, \cite{UKFuel2021} and \cite{AUFuel2021}. A consideration to take into account is that there may be many entries in the databases for a same make, model, year, fuel type and route type. In this cases, we use as catalog reference the median fuel value over all those entries.

\section{Evaluation}
We use the XAI proposal described in Subsection 3.2 different industry data sets to evaluate the hypotheses described below. For all these hypotheses we use as source of information the feature-relevance explanations yielded by the XAI algorithm, since they account for the individual impact from the different features in the fuel consumption. The hypotheses are connected to the evaluation steps mentioned before in Figure \ref{figure:SchemaProcessFuelCompare}.

Regarding the "Model Evaluation" step:
\begin{itemize}
    \item \textbf{Hypothesis 1 (H1):} The model's performance from the XAI algorithm in terms of the median Mean Average Precision Error (MAPE) and Adjusted R2 is good enough according to the references from the literature on each data set. 
\end{itemize}
Regarding H1, we include a comparison against other blackbox models, as well as a comparison against simpler whitebox ones. Particularly, we include a Linear Regression model with the usage of the ElasticNet \cite{zou2005regularization} algorithm as the whitebox alternative, and for black box models, we use the following tree based methods: XGBoost \cite{chen2016xgboost} and LightGBM \cite{ke2017lightgbm}.

For the "Domain Knowledge Evaluation" (Evaluation 1) step:
\begin{itemize}
    \item \textbf{Hypothesis 2 (H2):} The relative fuel impact explained for the different feature subclasses from Tables \ref{table:FARUsedpt1} and \ref{table:FARUsedpt2} is between the literature limits shown in \cite{zacharof2016review}, or at least below the maximum limit.
\end{itemize}

As for the "Domain Knowledge Evaluation" (Evaluation 2) step:
\begin{itemize}
    \item \textbf{Hypothesis 3 (H3):} The relative fuel impact explained for the vehicle-dates with anomalous fuel consumption is not significantly lower than the relative extra fuel detected by the outlier detection algorithm.
    \item \textbf{Hypothesis 4 (H4):} The new average fuel consumption after applying the recommendations from the "Daily Explanations" step is similar to both the catalog reference for that same vehicle's make, model, year, fuel type and on that route type. It will aso be similar compared to the median historical value of the vehicle's from the same group without fuel anomalies. This will be measured in terms of the MAPE against those reference values. 
\end{itemize}

The reference values for MAPE in H1 and H5 are the ones that appear in \cite{lewis1982industrial}. Those reference values are originally expressed for forecasting models, but we will use them for this use-case of regression, since MAPE is a metric also used for regression models \cite{de2016mean}.
\begin{itemize}
    \item $<10  $: Highly accurate forecasting
    \item $10-20$: Good forecasting
    \item $20-50$: Reasonable forecasting
    \item $>50$:   Inaccurate forecasting
\end{itemize}

Adjusted R2 indicates the proportion of the variance in the target feature that can be predicted using the input features. Even though this metric significantly depends on the context and units of the target feature \cite{hair2016primer, hair2013partial}, making if difficult to find general threshold values that indicate if it is good or not, there are some guidelines that may be considered. One of these guidelines is the proposal of \cite{chin1999structural}, that mentions the following thresholds:

\begin{itemize}
    \item $0.67$: Substantial
    \item $0.33$: Moderate
    \item $0.19$: Weak
\end{itemize}

\subsection{data sets}
The data source is the real-time feed of data retrieved by a telematics device connected to the to the on-board diagnostics (OBD) port. Particularly, we retrieve the data from devices connected to OBD-II port, since they allow an easy and direct retrieval of relevant features, such as the fuel consumption through the Engine Fuel Rate with the Parameter ID (PID) 015E \cite{ISO14230}. A sample of these raw data with a csv structure can be seen in (Table \ref{table:sample_structure}), and is also available at \cite{xaifuelabg2020}.

\begin{table}[]
\centering
\resizebox{\textwidth/2}{!}{%
\begin{tabular}{@{}llll@{}}
\toprule
\textbf{time\_tx} & \textbf{vehicle\_id} & \textbf{variable\_id} & \textbf{variable\_value} \\ \midrule
2020-10-31 00:02:34.073000+00:00 & b123 & EngineSpeed  & 1200 \\
2020-10-31 00:12:34.073000+00:00 & b124 & VehicleSpeed & 55   \\
2020-10-31 01:12:34.073000+00:00 & b125 & EngineSpeed  & 1200 \\
2020-10-31 02:02:34.073000+00:00 & b124 & TripFuel     & 3.1  \\ \bottomrule
\end{tabular}%
}
\caption{Sample of the received data from the IoT devices}
\label{table:sample_structure}
\end{table}

This data feed is then aggregated with a daily frequency into daily aggregations, building what we call the \textit{Fleet Analytics Record} (FAR), a database for each fleet owner where there is a record for every vehicle and date with different features related to the fuel consumption, as indicated in Tables \ref{table:FARUsedpt1} and \ref{table:FARUsedpt2}. It also includes other features related to the vehicle itself, such as the Vehicle Identification Number (VIN), make, model, manufacturing year or the fuel type.
Over the FAR, we apply several quality assurance analysis, where we discard records that are not meaningful or that may not be useful for training the model. The criteria followed is:
\begin{itemize}
    \item Remove records with missing trip distance, or with a low trip distance in that day (Km < 5).
    \item Remove records with missing fuel, or with a too small fuel consumption, according to Equation \ref{eq:anomalies-fuel}. This Equation is used first for discarding vehicles with an average fuel consumption that is too low (below the lower limit). Then, after the Noise Elimination step, it will be applied again with the remaining data for obtaining the upper limit for the univariate anomaly detection step from Figure \ref{figure:SchemaProcessFuelCompare}.
    \item Remove records with a potential wrong fuel value due to being extremely high. Besides having fuel anomalies that correspond to certain feature values, there is also noisy data regarding fuel consumption that needs to be eliminated in order to help the training of the ML model. We also use Equation \ref{eq:anomalies-fuel}. This data points are also removed before computing the final limits for the anomalous fuel consumptions that are not noisy data at the univariate anomaly detection step from Figure \ref{figure:SchemaProcessFuelCompare}.
\end{itemize}

Finally, we fill the other missing values with the median historical value from the fleet using vehicles from the same make, model, year and fuel type to fill the missing feature value of a particular vehicle on a particular date.

\begin{figure}[]
\centering
 \begin{tabular}{c@{\qquad}c@{\qquad}c}
\includegraphics[width=0.6 \columnwidth]{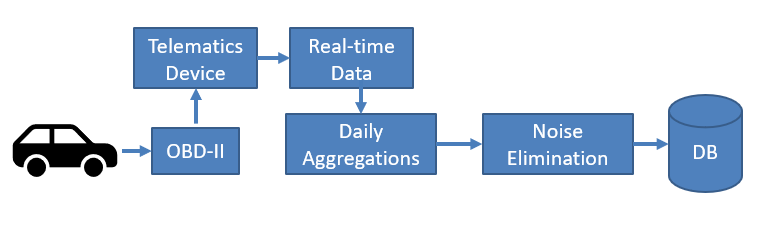}
  \end{tabular} 
  \caption{Data feed process for obtaining the databases.  \label{figure:DataFeedOBDII}}
\end{figure}

This process is done for different fleets independently, resulting in an individual data set for each one of them. With that, we use 9 data sets from different fleets, as indicated in Table \ref{table:data-description}. These data sets contain different types of vehicles that are identified with two groups of variables. The first one is the vehicle's make, model, year and fuel type. Since fuel consumption depends on the type of vehicle (among other things), we use the Vehicle's Identification Number (VIN) to identify those variables. With that, we get the different types of models that appear in column "N models". 
Along with that, since some models may have similar fuel consumption, we add an additional variable, named vehicle class, that groups together those vehicles (e.g. "Large Pick-Ups"). This vehicle class is inferred directly from the historical mean average fuel consumption, following the classification Table \ref{table:vehicle-class} that appears in \cite[p.~18]{national2010technologies}.
With that, we are conducting the analyses over fleets of vehicles that are different among themselves, in order to provide results that are as general as possible. Those fleets include passenger fleets (such as D1) of vehicles, as well as heavy-duty vehicles like trucks (such as D3). We are also covering different fleet sizes, such as "Large/Enterprise" (D1 and D2), "Medium" (D3, D4, D5 and D6), and "Small" (D7, D8 and D9), following the categorization of \cite{fleet2021trends}, where fleets with more than 500 vehicles are considered "Large/Enterprise"), fleets between 50 and 499 "Medium", and fleets with less than 49 vehicles "Small". This is indicated in column "Size".
Finally, column "N outliers" indicate the vehicle-dates with fuel anomalies, according to the univariate outlier detection from Section 3.

\begin{table}[]
\centering
\resizebox{\textwidth}{!}{%
\begin{tabular}{@{}lllllllllllllllll@{}}
\toprule
\textbf{Fleet} &
  \textbf{N vehicles} &
  \textbf{N models} &
  \textbf{N points} &
  \textbf{N outliers} &
  \textbf{Size} &
  \textbf{Class 0} &
  \textbf{Class 1} &
  \textbf{Class 2} &
  \textbf{Class 3} &
  \textbf{Class 4} &
  \textbf{Class 5} &
  \textbf{Class 6} &
  \textbf{Class 7} &
  \textbf{Class 8} &
  \textbf{Class 9}  &
  \textbf{Class 10} \\ \midrule
D1 & 1551 & 16 & 218038 & 5770  & Large  & 1479 & 34  & 1   & 37  & 0  & 0 & 0  & 0 & 0  & 0  & 0  \\
D2 & 1564 & 16 & 120605 & 1809  & Large  & 201  & 697 & 75  & 588 & 0  & 3 & 0  & 0 & 0  & 0  & 0  \\
D3 & 316  & 44 & 65285  & 10475 & Medium & 243  & 5   & 1   & 9   & 10 & 5 & 13 & 9 & 21 & 0  & 0  \\
D4 & 252  & 14 & 35283  & 1915  & Medium & 4    & 178 & 61  & 9   & 0  & 0 & 0  & 0 & 0  & 0  & 0  \\
D5 & 165  & 20 & 22402  & 724   & Medium & 165  & 0   & 0   & 0   & 0  & 0 & 0  & 0 & 0  & 0  & 0  \\
D6 & 143  & 20 & 18555  & 2002  & Medium & 1    & 28  & 100 & 5   & 3  & 0 & 0  & 0 & 5  & 1  & 0  \\
D7 & 33   & 5  & 8450   & 942   & Small  & 0    & 0   & 0   & 2   & 3  & 0 & 0  & 0 & 10 & 18 & 0  \\
D8 & 20   & 5  & 2125   & 349   & Small  & 5    & 15  & 0   & 0   & 0  & 0 & 0  & 0 & 0  & 0  & 0  \\
D9 & 3    & 2  & 269    & 10    & Small  & 1    & 0   & 0   & 0   & 0  & 0 & 0  & 0 & 2  & 0  & 0  \\ \bottomrule
\end{tabular}%
}
\caption{data set description, including the number of datapoints, number and type of vehicles.}
\label{table:data-description}
\end{table}

\begin{table}[]
\centering
\resizebox{\textwidth}{!}{%
\begin{tabular}{@{}lllllll@{}}
\toprule
\textbf{Class} &
  \textbf{Applications} &
  \textbf{Gross Weight Range (lb)} &
  \textbf{L100Km\_min} &
  \textbf{L100Km\_max} &
  \textbf{L100Km\_med} &
  \textbf{Vehicle Class} \\ \midrule
1c & Cars only                                                                       & (3200)-6000   & 7.12  & 9.41            & 8.27  & 0  \\
1t & Minivans, Small SUVs, Small Pick-Ups                                            & (4000)-6000   & 9.40  & 11.76          & 10.58 & 1  \\
2a & Large SUVs, Standard Pick-Ups                                                   & 6001-8500     & 11.20 & 11.76         & 11.48 & 2  \\
2b & Large Pick-Up, Utility Van, Multi-Purpose, Mini-Bus, Step Van                   & 8501-10,000   & 15.68 & 23.52         & 19.60 & 3  \\
3  & Utility Van, Multi-Purpose, Mini-Bus, Step Van                                  & 10,001-14,000 & 18.09 & 29.40       & 23.74 & 4  \\
4  & City Delivery, Parcel Delivery, Large Walk-in, Bucket, Landscaping              & 14,001-16,000 & 19.60 & 33.60 & 26.60 & 5  \\
5  & City Delivery, Parcel Delivery, Large Walk-in, Bucket                           & 16,001-19,500 & 19.60 & 39.20           & 29.40 & 6  \\
6  & City Delivery, School Bus, Large Walk-in, Bucket                                & 19,501-26,000 & 19.60 & 47.04            & 33.32 & 7  \\
7 &
  City Bus, Furniture, Refrigerated, Refuse, Fuel Tanker, Dump,Tow, Concrete,Fire Engine,Tractor-Trailer &
  26,001-33,000 &
  29.40 &
  58.80 &
  44.10 &
  8 \\
8b & Tractor-Trailer: Van, Refrigerated, Bulk Tanker, Flat Bed (combination trucks)  & 33,001-80,000 & 31.36 & 58.80        & 45.08 & 9  \\
8a & Dump, Refuse, Concrete, Furniture, City Bus, Tow, Fire Engine (straight trucks) & 33,001-80,000 & 39.20 & 94.09            & 66.64 & 10 \\ \bottomrule
\end{tabular}%
}
\caption{Vehicle classes according to their average fuel consumption, as appears in \cite[p.~18]{national2010technologies} }
\label{table:vehicle-class}
\end{table}

As already mentioned, the features considered for each of those data sets appear in Tables \ref{table:FARUsedpt1} and \ref{table:FARUsedpt2}. Column "Name" includes a descriptive name for each of the features, and column "Description" contains a descriptive text about each of them. "Unit" indicates the metric units associated to each of the features, and "Notes" contains a description about some of the variables and why they may impact in fuel consumption (particularly for the ones that are not trivial). The column "Type" shows the type of impact that those features have in fuel consumption. If the type is "Positive" it indicates that increasing that feature value will normally \textit{increase} fuel usage. An example of this is the number of events with high RPM; more events lead to more fuel consumption. On the contrary, if the type is "Negative", it indicates that increasing that feature value will normally \textit{decrease} fuel usage. An example of this is the time using speed control; more time using it should lower the fuel consumption (versus not using it). Another example is the tire pressure; when they decrease, the fuel used will increase. Column "Reference Zero" indicates the columns that in order to see the impact in the fuel consumption are set to zero. For instance, for obtaining the feature impact for a variable like "rpm\_high", this variable is set to 0 for calculating the reduction in the fuel consumption due to it by seeing the decrease with respect to the current feature value. For the remaining features the reference is, by default, the median value for that feature over the vehicles with fuel inliers from the same vehicle model.
Finally, columns "Category" and "Subcategory" refer directly to the same columns from Table \ref{table:FeatureInfluenceReduced} from \cite{zacharof2016review}. The columns that do not have a value in both of these columns are columns that are not features used for explaining the fuel (they are relevant for the data set, and some of them are even used in the model, like the vehicle model, but they are not used for explanations). Among these columns is the main driving context detected for each day ("route\_type"). This is calculated as follows:

\begin{itemize}
    \item IF $per\_time\_city \leq low\_th\_time$ AND $trip\_kms \geq th\_kms$ THEN $route\_type = hwy$
    \item ELSE IF $per\_time\_city \geq high\_th\_time$ AND $trip\_kms \leq th\_kms$ THEN $route\_type = city$
    \item ELSE $route\_type = combined$
\end{itemize}

\begin{figure}[]
\centering
 \begin{tabular}{c@{\qquad}c@{\qquad}c}
\includegraphics[width=0.6 \columnwidth]{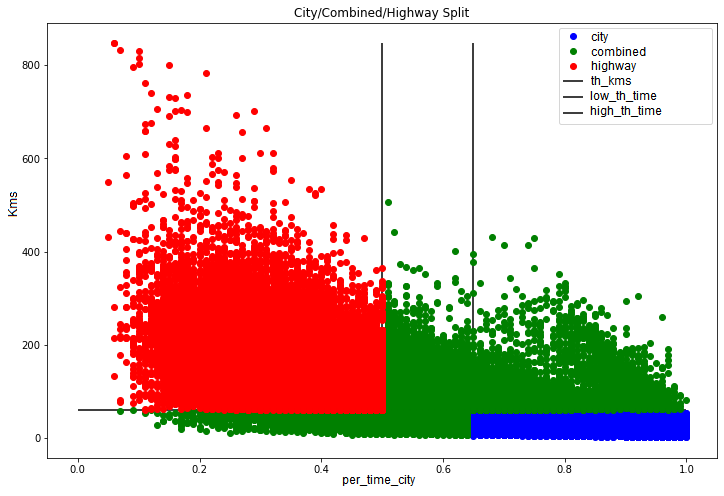}
  \end{tabular} 
  \caption{Daily categorization of route types based on the trip distance (Km) and per\_time\_city for the data set D1.  \label{figure:split_city_comb_hwy}}
\end{figure}

With $th\_kms = 30$, $low\_th\_time = 0.5$ and $high\_th\_time = 0.65$.
Thus, we categorize each vehicle-date with a particular route type that may be "city", "highway" or "combined", depending on the total trip kms (trip\_kms) and the value of the variable per\_time\_city. An example of this route type categorization, using the threshold values aforementioned, appears in Figure \ref{figure:split_city_comb_hwy}.

The categories considered are "Auxilary Systems" (for all the features that imply an additional electrical energy consumption), "Driving Behaviour" (driver-related features), "Road Conditions", "Vehicle Conditions" and "Weather Conditions". Regarding "Vehicle Conditions", we have included additional variables within the "Other" subcategory with respect to \cite{zacharof2016review} (e.g. the additional fuel consumption when the DEF level is low), so it does not match the ones covered in that review. Because of that, this subcategory will not be used for checking the hypotheses already mentioned. For "Rain" subcategory, since the review only provides one reference value, we will only check H1 (since there are no limits per se).

Finally, the dependent variable is the average fuel consumption, calculated as follows:
\begin{equation}
    avg\_fuel\_consumption =  \frac{trip\_fuel\_used}{trip\_kms} \times 100
\end{equation}

\subsection{Results}
We divide the Results Subsection into three parts: first, regarding the model evaluation analysis, and second, regarding the evaluation of explanations within the context of the domain knowledge. We end with an analysis of the potential impact of the explanations for both fuel saving and emissions reduction.

\subsubsection{Model Evaluation}
Considering a train/test split for each data set of 90/10, we get the results shown in Table \ref{table:MetricsModelAll}. As we can see, D1, D2, D4, D5, D6, D7 and D8 are within the "highly accurate forecasting" category, while D3 and D9 are within the "good forecasting" one. 
Thus, the model is able to infer sufficiently good relationships between the input data and the fuel consumption, and H1 is validated. Because of that, it can be used for extracting explanations in order to evaluate the remaining hypotheses. In fact, the explanations for all those fleets over that 4 months period yielded explanations for 74971 vehicle-dates. When considering vehicles with a median MAPE value over the test set if "Good forecasting" or better, the explanations covered corresponds 96\%, and 76\% when considering vehicles with a "Highly accurate forecasting".
Complementing this, in Table \ref{table:MetricsModelAll} we see the adjusted R2 value on the test set for each data set. Every data set is within the "substantial" R2 category, with the exception of D4 and D8.
Compared to the other models, we see how EBM improves the ElasticNet model for every metric and in every data set. Regarding LightGBM and XGBoost, EBM is similar for most fleets and for both metrics. In some cases, like D7, EBM is actually better than the other two models for both metrics. In D1, EBM is better in terms of MAPE than LightGBM.

\begin{table}[]
\centering
\resizebox{325pt}{!}{%
\begin{tabular}{@{}lcccccccc@{}}
\toprule
\textbf{} & \multicolumn{4}{c}{\textbf{MAPE}}           & \multicolumn{4}{c}{\textbf{Adjusted R2}}     \\
\textbf{Fleet} & \textbf{Linear Model} & \textbf{LightGBM} & \textbf{XGBoost} & \textbf{EBM} & \textbf{Linear Model} & \textbf{LightGBM} & \textbf{XGBoost} & \textbf{EBM} \\ \midrule
D1        & 17 & 9           & \textbf{8}  & \textbf{8} & 21 & 77          & \textbf{79} & 77          \\
D2        & 17 & \textbf{8}  & \textbf{8}  & \textbf{8} & 46 & 84          & \textbf{86} & 82          \\
D3        & 28 & 11          & \textbf{10} & 13         & 78 & 95          & \textbf{96} & 94          \\
D4        & 14 & 9           & \textbf{8}  & 9          & 16 & 67          & \textbf{72} & 61          \\
D5        & 15 & 9           & \textbf{8}  & 9          & 13 & 71          & \textbf{73} & 66          \\
D6        & 15 & \textbf{7}  & \textbf{7}  & 9          & 51 & 87          & \textbf{90} & 85          \\
D7        & 32 & 23          & 22          & \textbf{8} & 44 & 67          & 67          & \textbf{80} \\
D8        & 14 & \textbf{8}  & \textbf{8}  & \textbf{8} & 7  & 62          & \textbf{64} & 63          \\
D9        & 17 & \textbf{12} & 14          & 15         & 77 & \textbf{88} & 87          & 85          \\ \bottomrule
\end{tabular}%
}
\caption{MAPE median results and adjusted R2 for each fleet.}
\label{table:MetricsModelAll}
\end{table}

\subsubsection{Domain Knowledge Evaluation}
\begin{figure}[]
\centering
 \begin{tabular}{c@{\qquad}c@{\qquad}c}
\includegraphics[width=0.95 \columnwidth]{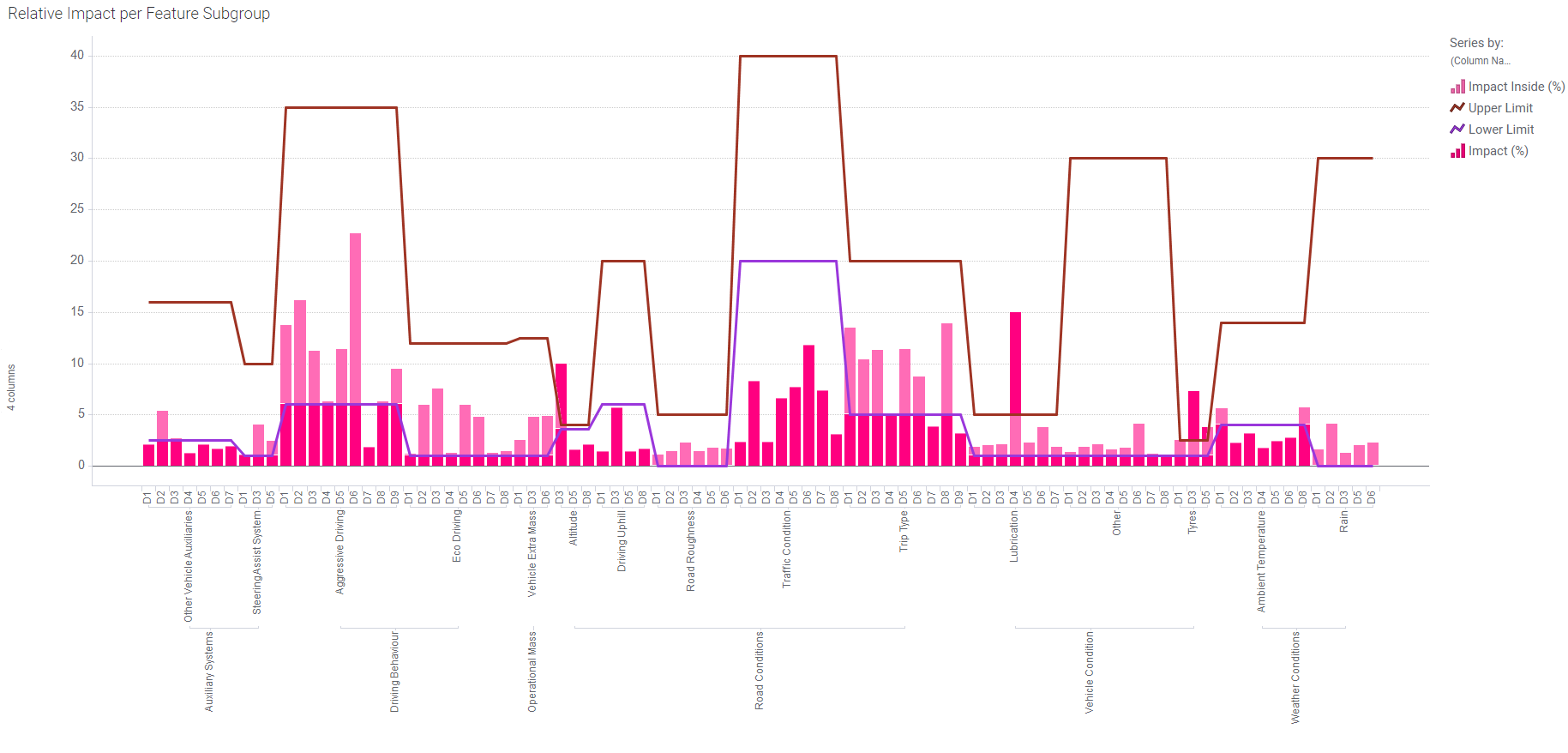}
  \end{tabular} 
  \caption{Median feature impact per Category-Subcategory-Fleet and the corresponding limits from the literature \cite{zacharof2016review}  \label{figure:FeatureImpactperdata set}}
\end{figure}

\begin{figure}[]
\centering
 \begin{tabular}{c@{\qquad}c@{\qquad}c}
\includegraphics[width=0.95 \columnwidth]{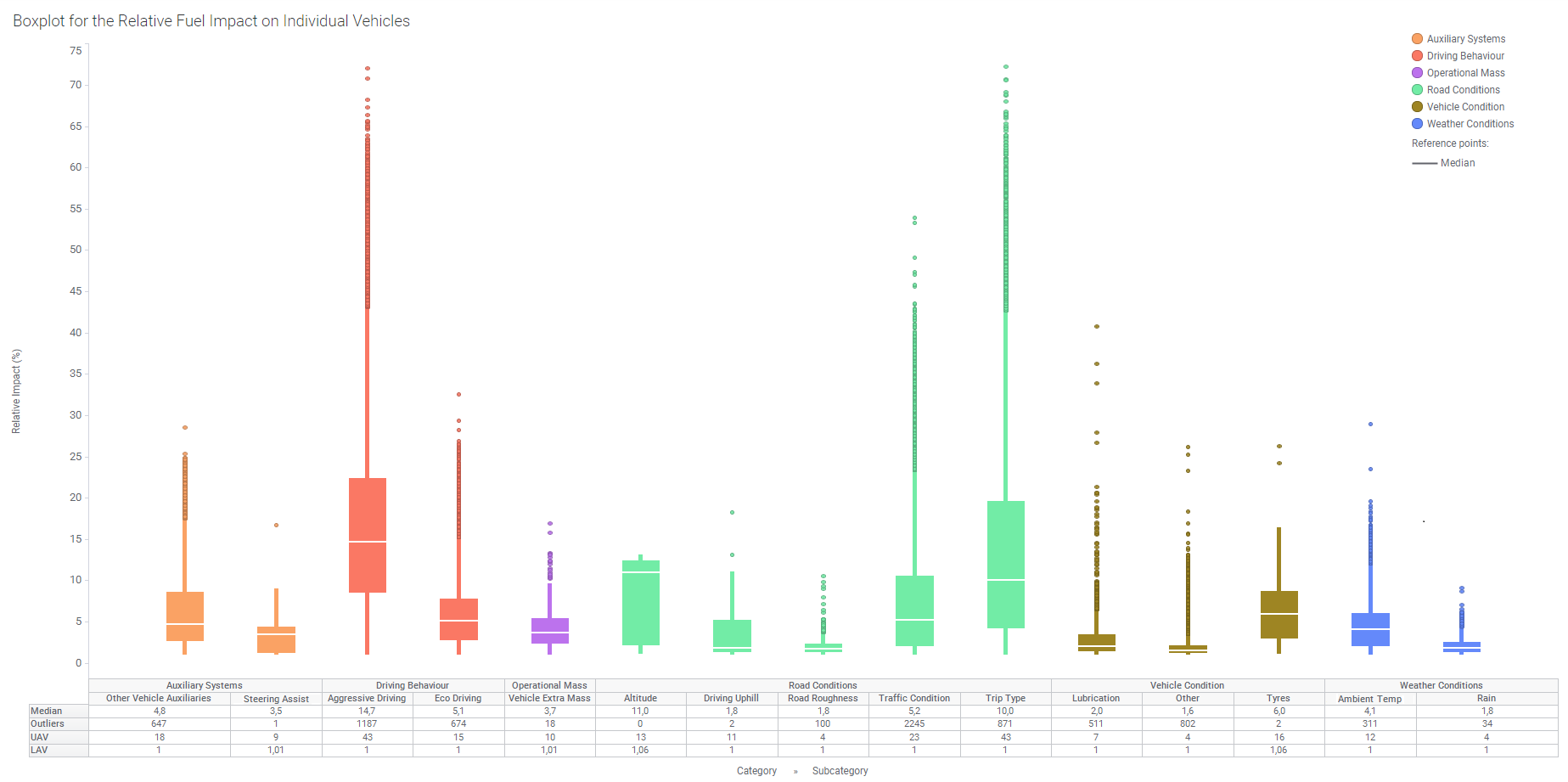}
  \end{tabular} 
  \captionsetup{justification=centering}
  \caption{Fuel impact per vehicle-date for each fuel factor subcategory.  \label{figure:BoxPlotVehicleFuelImpact}}
\end{figure}

For evaluating the explanations, we focus on the 4 months of data where the winter period is included (in order to be able to assess the impact of the ambient temperature). Using the models, we get the explanations for each vehicle-date for that period of data, and we aggregate the median feature impact values per subcategory and per vehicle fleet. The median results regardless of the fleet appear in Figure \ref{figure:BoxPlotVehicleFuelImpact}, and the median results considering fleet and including the limits from the SOTA appear in Figure \ref{figure:FeatureImpactperdata set}. For the analyses, we have considered only the vehicles that have a median MAPE over the test set of "Good forecasting" or better (unless otherwise indicated).

In Figure \ref{figure:BoxPlotVehicleFuelImpact} we see how the absolute relative impact for each subcategory is below the limits show in Table \ref{table:FeatureInfluenceReduced}. This is shown clearer in Figure \ref{figure:FeatureImpactperdata set}, where we see that for all the data sets and for all the feature subcategories the relative impact is below the reference values. 

Complementing this, in Table \ref{table:FeatureInfluenceReduced} we see what combinations of subcategories-data sets are within the limits from the literature. For 44 combinations, out of the 77 (without the Subcategories of "Other" and "Rain", as mentioned before), the feature relevance is within the limits from the SOTA. The remaining 33 that are not within the limits is because they are either lower than the minimum value used, or higher (for Tyres and Altitude in D3, Lubrication in D4 and Tyres in D5). "Aggressive Driving", "Eco-Driving", "Trip Type" and "Road Roughness" are the Subcategories that are both common in all data sets while having an aggregated feature impact that is within the literature limits. Others, such as "Steering Assist Systems" and "Vehicle Extra Mass" are also fully within the limits, but they are features that are relevant only for some data sets.

With Figure \ref{figure:BoxPlotVehicleFuelImpact} we see the individual impact per vehicle and date, for all the data sets considered together. As the Figure shows, "Other Vehicle Auxiliaries", "Steering Assist Systems", "Aggressive Driving", "Eco-Driving", "Vehicle Extra Mass", "Lubrication", "Road Roughness", "Trip Type" and "Ambient Temperature" have a median value per vehicle-date that is within the limits from the SOTA. For some Subcategories, such as "Steering Assist Systems", "Vehicle Extra Mass", "Driving Uphill", "Road Roughness", "Traffic Condition", and "Ambient Temperature", the upper whisker value from the boxplot is also within the SOTA limits. In general, all the median values are at least below the upper limits, with the exception on "Altitude" and "Tyres", which are overestimated by the model.
We also see that even though the impact per Subcategory normally does not exceed the upper values reported, there are data points where the impact is above the thresholds from the literature. 

With that, H2 is validated for the 73 out of the 77 combinations of subcategories-data sets since they have an influence in the fuel consumption because the relative impact is always below the maximum SOTA values, and in some cases, its even between them. 

\begin{figure}[]
\centering
 \begin{tabular}{c@{\qquad}c@{\qquad}c}
\includegraphics[width=0.9 \columnwidth]{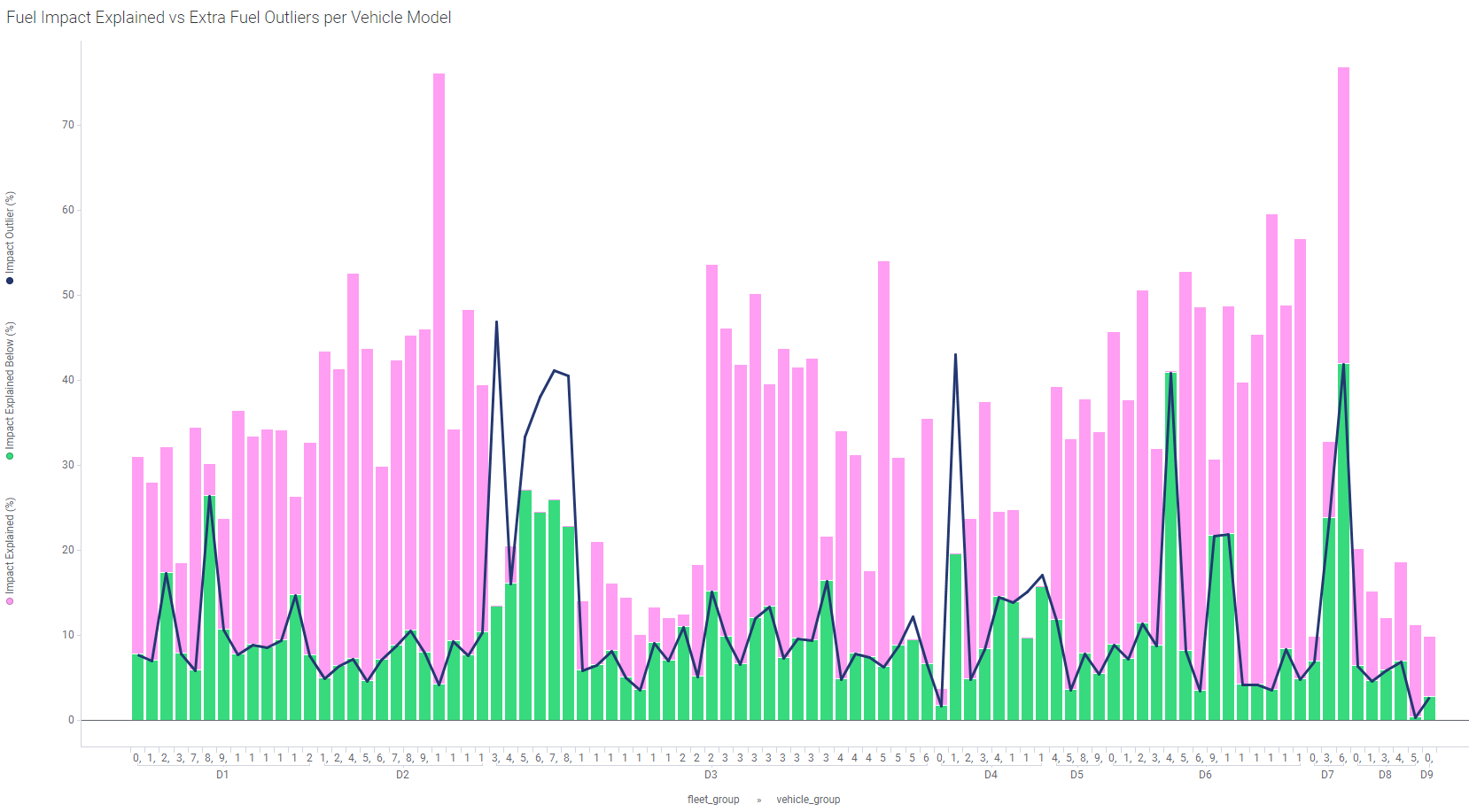}
  \end{tabular} 
  \captionsetup{justification=centering}
  \caption{Relative extra fuel explained versus the relative extra fuel by the outlier detection method for each of the vehicle models within every data set. \label{figure:ImpactExplainedperOutliersModel}}
\end{figure}

For H3, we analyse the extra fuel explained through XAI with respect to the extra fuel indicated by the limits generated by the outlier detection method from the previous Section. The intuition behind it is that even though the XAI whitebox algorithm is not trained over all the potential causes that may impact in the fuel consumption, it is enough to explain at least that extra anomalous fuel. In Figure \ref{figure:ImpactExplainedperOutliersModel} we see that the comparison between the relative extra fuel explained by the XAI method versus the extra fuel shown by the outlier detection algorithm for each of the models within every fleet data set. We see that, in fact, for the majority of the cases the extra fuel explained is actually superior than the extra anomalous fuel detected. Table \ref{table:FuelExplainedvsOutliers} shows that comparison for each data set and for every vehicle and date. It shows that D4, D5, D7, D8 and D9 are the only data sets where there are no significant differences between anomalous fuel and explained extra fuel. In all the remaining cases, the explained fuel is actually superior to the outlier part. So, the XAI method is actually able to explain at least the anomalous part of the fuel in all the cases, validating H3.

\begin{table}[]
\centering
\resizebox{175pt}{!}{%
\begin{tabular}{@{}llll@{}}
\toprule
\textbf{Fleet} & \textbf{Median (explained)} & \textbf{Median (outliers)} & \textbf{p-value} \\ \midrule
D1 & \textbf{0.31} & 0.09 & \textless 0.01  \\
D2 & \textbf{0.44} & 0.07 & \textless 0.01  \\
D3 & \textbf{0.24} & 0.09 & \textless 0.01  \\
D4 & 0.22 & 0.14 & 0.21            \\
D5 & 0.36 & 0.07 & 0.02            \\
D6 & \textbf{0.47} & 0.08 & \textless 0.01  \\
D7 & 0.10 & 0.7 &  0.32            \\
D8 & \textbf{0.15} & 0.06 & 0.01            \\
D9 & 0.11 & 0.03 & 0.32            \\ \bottomrule
\end{tabular}%
}
\captionsetup{justification=centering}
\caption{Relative impact of extra fuel explained versus extra fuel with respect of the outlier limit.}
\label{table:FuelExplainedvsOutliers}
\end{table}

In order to check H4, we get the MAPE metric from the new average fuel consumption for each vehicle-date after applying the recommendations, and compare it to the catalog fuel consumption for the same make, model, year, fuel type and route type, as mentioned in Subsection 3.3. The same procedure is applied for comparing it against the median fuel consumption from the vehicles-dates data points corresponding to the same group, and on the same route type, that are not identified as an anomalous fuel consumption. Results appear in Table \ref{table:NewFuelVSCatalogAndMedian}. MAPE 1 corresponds to the median MAPE versus the catalog fuel, MAPE 2 corresponds to the median MAPE versus the median fuel inlier vehicles, and MAPE 3 is the same as MAPE 2 but considering only explanations for outlier vehicles. 
Considering MAPE 1, we see that for every data set, except for D3 and D9, the new average fuel consumptions after applying the recommendations are similar to the catalog fuel ("Reasonable forecasting" or better). D6 does not have any results in the catalog comparisons because the make, models and fuel types did not appear within the databases used in Subsection 3.3.
Columns "\% MAPE 1 < 0.5", "\% MAPE 1 < 0.2" and "\% MAPE 1 < 0.1" show the percentage of vehicle-date explanations that have an individual MAPE of 0.5, 0.2 or 0.1 respectively. 

Regarding the comparison against the median inliers, the results are better. All the data sets yield a median MAPE from the "Reasonable forecasting" category or better, except for D9. There are significant differences in some cases, specially for D9, meaning that the fuel consumption of the inliers is different from the catalog (in the case of D9, the data set is small, so the anomaly detection process may not yield accurate results). For other cases, like D1, the results are very similar (indicating that the reference fuel is indeed similar to the median inlier consumption for those vehicle models).

The results either improve or are similar when considering only the fuel outliers in the explanation process. In this case, the worst result is D7, with a MAPE of 0.41. This indicates that even though the fuel explained in D7 is sufficient to cover the outlier part, is not enough to cover all the extra fuel until reaching the median inlier value.
The rest of the data sets are either in the "Good forecasting" category or close to it.

Finally, the column "\% below catalog" show the percentage of vehicle-dates that are receiving a recommendation that turns the average fuel consumption (L/100Km) below the catalog reference (with an offset of 1 L/100Km). This metric should be minimized, because the target fuel should not be below the catalog reference (is a value that is not physically reachable). This indicates data points that the model is nor explaining properly (its overestimating the potential fuel reduction). The best cases are D1, D4, D7 and D9 where this metric is 3.3\%, 1.9\%, 4.0\% and 0\% respectively.

\begin{table}[]
\centering
\resizebox{\textwidth}{!}{%
\begin{tabular}{lllllllllll}
\hline
\textbf{Fleet} &
  \textbf{MAPE 1} &
  \textbf{MAPE 2} &
  \textbf{MAPE 3} &
  \textbf{\begin{tabular}[c]{@{}l@{}}\% MAPE 1\\  \textless 0.5\end{tabular}} &
  \textbf{\begin{tabular}[c]{@{}l@{}}\% MAPE 1\\  \textless 0.2\end{tabular}} &
  \textbf{\begin{tabular}[c]{@{}l@{}}\% MAPE 1\\  \textless 0.1\end{tabular}} &
  \textbf{\begin{tabular}[c]{@{}l@{}}\% MAPE 2\\  \textless 0.5\end{tabular}} &
  \textbf{\begin{tabular}[c]{@{}l@{}}\% MAPE 2\\  \textless 0.2\end{tabular}} &
  \textbf{\begin{tabular}[c]{@{}l@{}}\% MAPE 2\\  \textless 0.1\end{tabular}} &
  \textbf{\begin{tabular}[c]{@{}l@{}}\% below \\ catalog\end{tabular}} \\ \hline
D1 & 0.16 & 0.13 & 0.21 & 90.0 & 58.8 & 33.1 & 97.0  & 70.3  & 41.8 & 3.3     \\
D2 & 0.47 & 0.34 & 0.24 & 24.0 & 6.9  & 3.5  & 84.0  & 19.4  & 8.5  & 34.0    \\
D3 & 0.52 & 0.30 & 0.32 & 30.8 & 7.5  & 4.1  & 81.8  & 33.1  & 18.7 & 39.4    \\
D4 & 0.34 & 0.11 & 0.19 & 76.2 & 22.5 & 7.0  & 99.0  & 83.4  & 47.2 & 1.9     \\
D5 & 0.37 & 0.27 & 0.17 & 29.0 & 7.3  & 2.7  & 94.6  & 28.1  & 10.9 & 30.6    \\
D6 &      & 0.41 & 0.29 &      &      &      & 72.4  & 10.5  & 4.9  &         \\
D7 & 0.22 & 0.05 & 0.41 & 13.5 & 7.9  & 4.6  & 90.8  & 82.4  & 73.1 & 4.0     \\
D8 & 0.16 & 0.05 & 0.07 & 36.8 & 22.0 & 13.7 & 99.6  & 93.5  & 78.0 & 13.0    \\
D9 & 1.06 & 0.05 & 0.20 & 0    & 0    & 0    & 100.0 & 93.2  & 70.5 & 0       \\ \bottomrule
\end{tabular}%
}
\caption{Different MAPE metrics on each of the data sets versus the catalog fuel consumption (MAPE 1), the median inliers (MAPE 2), or considering only the vehicles with outlier fuel consumption versus the inliers (MAPE 3).}
\label{table:NewFuelVSCatalogAndMedian}
\end{table}

\begin{figure}[]
\centering
 \begin{tabular}{c@{\qquad}c@{\qquad}c}
\includegraphics[width=395pt, height=580pt, keepaspectratio]{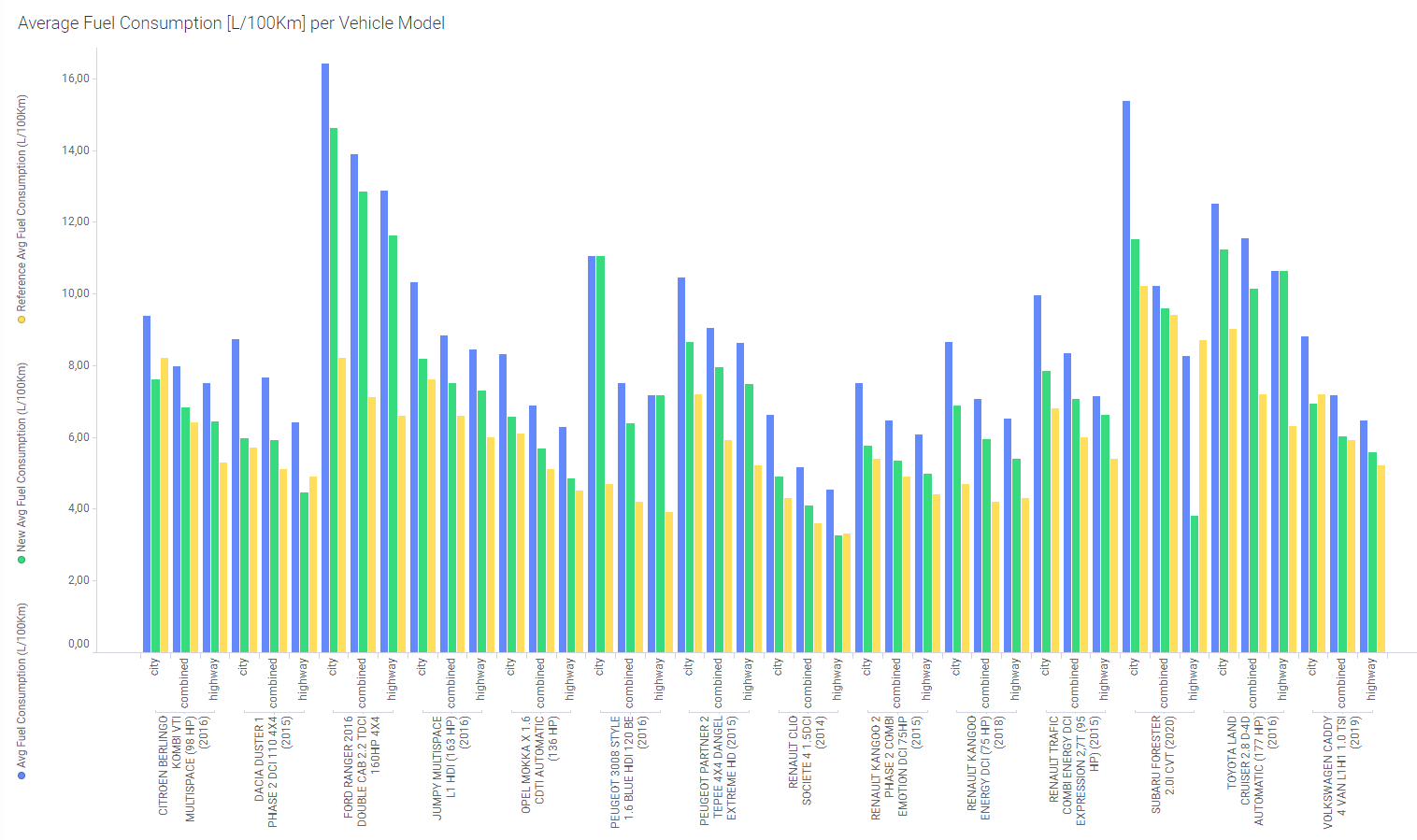}
  \end{tabular} 
  \caption{Average fuel consumption (L/100Km) for the vehicle's models from data set D1 in each route type (city, combined, highway), before and after applying the recommendations, and compared to the catalog reference. \label{table:ModelRefFuelD1}.}
\end{figure}

The previous analysis can be enhanced by checking the fuel reduction considering each vehicle model and route type with respect to the catalog reference. In the case of D1, we explicitly had the vehicle's makes (so it was not needed to retrieve them from the VIN decoding process). Because of that, we obtained there exactly catalog fuel consumption from \cite{motoreu2021} and \cite{ultimatespecs2021}. The results appear in Figure \ref{table:ModelRefFuelD1}. There, we only see 4 cases where the new average fuel is below the catalog reference.

\subsubsection{Potential Impact}
\begin{figure}[h!]
\centering
 \begin{tabular}{c@{\qquad}c@{\qquad}c}
\includegraphics[width=380pt, height=580pt, keepaspectratio]{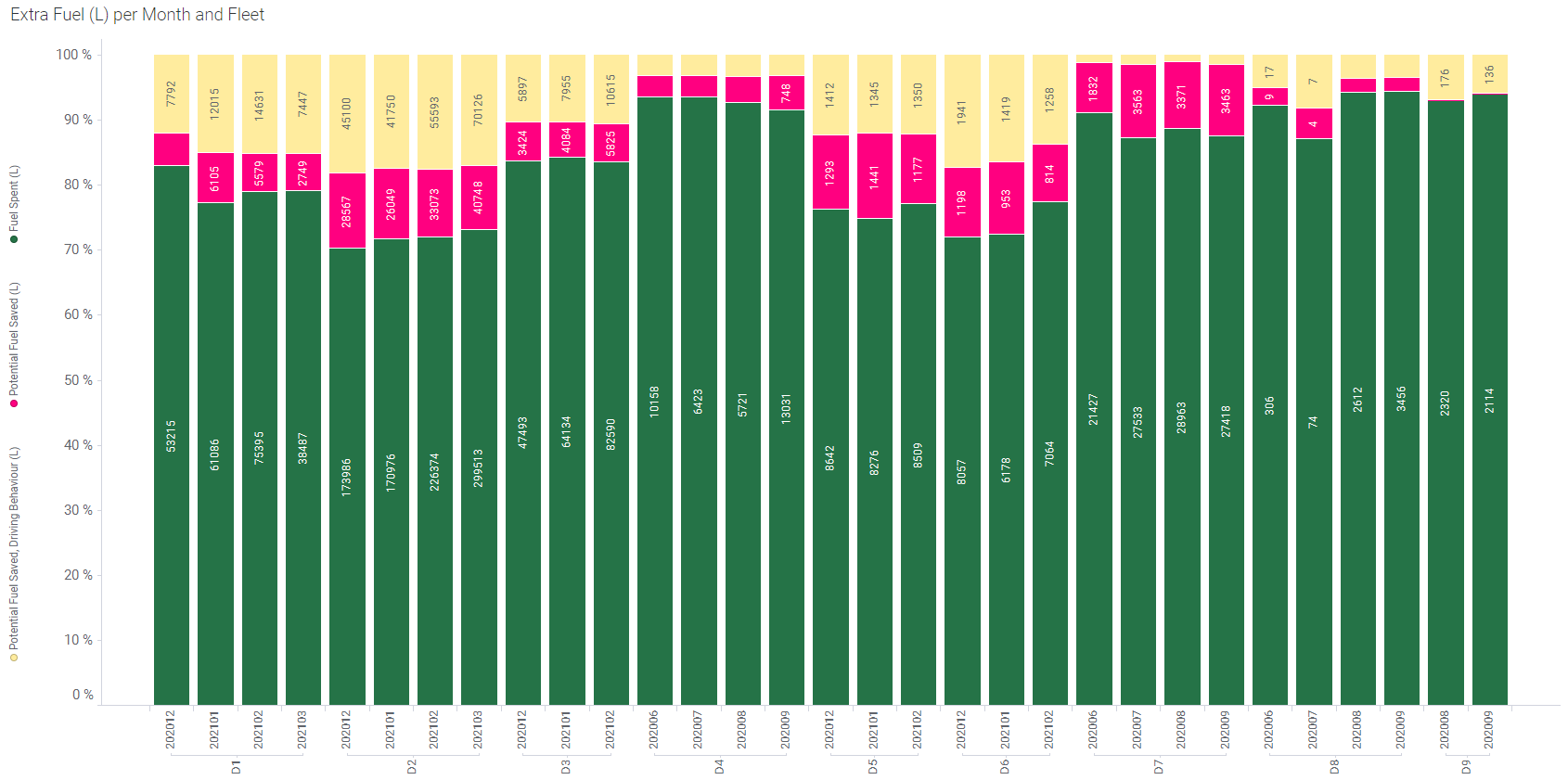}
  \end{tabular} 
  \caption{Monthly fuel consumption (L) for each fleet over the different months, along with the part of that fuel that corresponds to the extra fuel due to driving behaviour, together with the extra fuel from the remaining factors. \label{table:ExtraFuelPerMonth}}
\end{figure}

With the previous analyses we checked that EBM is indeed suitable for both model and explain the relationship between fuel factors and fuel consumption, since model metrics are good enough, and domain knowledge metrics are in general aligned with prior domain knowledge.
This was needed for conducting an analysis on the impact of fuel factors in order to quantify how much extra fuel is spent due to actionable features. 
For this analysis we focus on driving behaviour features, since they are among the features that have more impact. They are also actionable, because they are mainly associated to the drivers and can be potentially acted upon without changing other contextual factors, such as the planned routes. 
Figure \ref{table:ExtraFuelPerMonth} shows the fuel consumption on each of the fleets over the four months considered, together with the extra fuel consumption from driving behaviour, and the extra fuel consumption due to the remaining factors.

Focusing on D1 (since it is the fleet with more information and that provided better results), we see that the relative impact of all the features is between 17\% and 23\%, and for driving behaviour only, it is between 12\% and 15\%. Taking as an example the month of February, we see that there are 14631 extra litres spent due to driving behaviour. Reducing it would have a positive impact both in the expenses from the fleet, as well as in the environment. Since the vehicles from D1 are mostly diesel, using the conversion to CO2 from \cite{DieselLitresTOCo2}, where 2.67633 Kg of CO2 are emitted per liter of diesel spent, the extra CO2 emissions in one month due to driving behaviour is between 19930 and 39157 Kg.

\subsection{Software Used}
The main libraries used for the work done in this paper are the following:
\begin{itemize}
    \item EBM \cite{interpret_ml_repo}. We used the default parameters from the library for all the analyses.
    \item Hypothesis contrasts \cite{2020SciPy-NMeth}
    \item XGBoost \cite{xgboost_repo}
    \item LightGBM \cite{lightgbm_repo}
    \item ElasticNet \cite{scikit-learn}
\end{itemize}

\subsection{Limitations}
First of all, our proposal studies the influence on fuel usage for petrol and diesel vehicles altogether. The independent analysis for petrol and diesel vehicles may yield different results. Also, we do not cover hybrid vehicles within our study.
Though we focused on actionable features for analysing the impact on the fuel consumption through XAI, there are other features that could be elicited.
Regarding the EBM algorithm, we only used the individual feature relevance of each variable for building the recommendations, not considering possible pairwise terms if they exist.
Finally, as we saw for the case of harsh turns in Figure \ref{figure:ExpMethodExample}, the relationships between an input feature and output are not monotonic. It would be interesting to analyse how the results differ when applying monotonic constraints.

\subsection{Future Work}
We see several possible lines of work following this paper. First of all, the set of variables used could be enhanced. Even though we use up to 70 features per data set, not all the subcategories mentioned within the SOTA are covered. For instance, we do not use any feature that measures the usage of a trailer towing or roof racks. 
Also, the analyses could be complemented with the usage of other XAI feature relevance techniques for generalizing and comparing the results obtained.
Finally, even though the domain knowledge is considered before the model training and explanation generation when eliciting the features, for the remaining steps (business rules for filtering the explanations, metric analysis...) is something that is being applied post hoc. Applying all the knowledge before training the model could potentially yield better results. 

\section{Conclusion}
In this paper, we used Explainable Artificial Intelligence (XAI) through Explainable Boositng Machine (EBM) to measure the potential impact that actionable factors may have on the fuel consumption of diesel and petrol vehicles. EBM are whitebox Machine Learning (ML) models that have good model performance while having a good degree of interpretability.
To achieve that, we worked with real-world industry data sets that represent different types of vehicles, from passenger cars to heavy-duty trucks. We have gathered data from telematic devices connected to the vehicles for more than one year. With this source data, we have elicited up to 70 potential fuel factors based on the literature in order to build a data set that can be used for both model and explain the relationship between those factors and fuel consumption.

Then, after training an EBM model on each one of those data sets, we have proposed an algorithm to generate the explanations and quantify the potential impact of those fuel factors, while combining the explanations with business rules that contribute to align the explanations to prior domain knowledge.

After that, we have evaluated the quality of the model from two points of view. First, we have checked that the algorithm was able to properly model the relationship between input factors and fuel consumption analysing its model performance. At this point, we have also compared the performance against other well-known ML models with high predictive power that do not directly provide explanations (blackbox algorithms). We saw that EBM achieves good results that are similar to those other models. 
Second, we have evaluated the explanations generated versus prior domain knowledge from the State of the Art, in order to see if the explanations are meaningful and if they are aligned with that knowledge. We have seen that in general they are aligned, with particular good results for factors related driving behaviour, operational mass and trip type. 

With that, we proceeded to quantify the potential extra fuel that those vehicle fleets have across different months both in general and also specifically considering driving behaviour factors. For some of the vehicle fleets with more data, we saw potential impacts due to driving behaviour of around 15\%, highlighting both the extra economic costs and the environmental impact that the fleet is having due to inefficient driving.  

In this paper we followed the literature regarding factors that impact in the fuel consumption of a vehicle, and we used XAI in order to see if it is possible to explain and quantify that impact with these techniques. For that, we used Explainable Boosting Machine (EBM) algorithm, a type of whitebox Machine Learning model that yields explanations in terms of feature relevance. We trained the model with a set of up to 70 features in order to predict the fuel consumption of diesel and petrol vehicles, using several real-world industry data sets from very different types of fleets (passenger cars, trucks...). Then, we generated explanations combining the information provided by the EBM algorithm and the feature taxonomies from the SOTA regarding the factors that affect fuel consumption. 

\subsection{CRediT authorship contribution statement} 
\textbf{Alberto Barbado}: Conceptualization, Investigation, Writing - original draft, Writing - review and editing, Visualization, Formal Analysis, Methodology. 
\textbf{Óscar Corcho}: Writing - review and editing, Supervision.

\subsection{Acknowledgements} 
This research was done within the context of the registered patent \cite{patent2020fleet} for LUCA Fleet at Telefónica. We thank Pedro Antonio Alonso Baigorri, Federico Pérez Rosado, Raquel Crespo Crisenti and Daniel García Fernández for their collaboration.

\printendnotes

\bibliography{main}

\section{Annex}

\begin{table}[h!]
\centering
 \begin{tabular}{c@{\qquad}c@{\qquad}c}
\includegraphics[width=395pt, height=580pt, keepaspectratio]{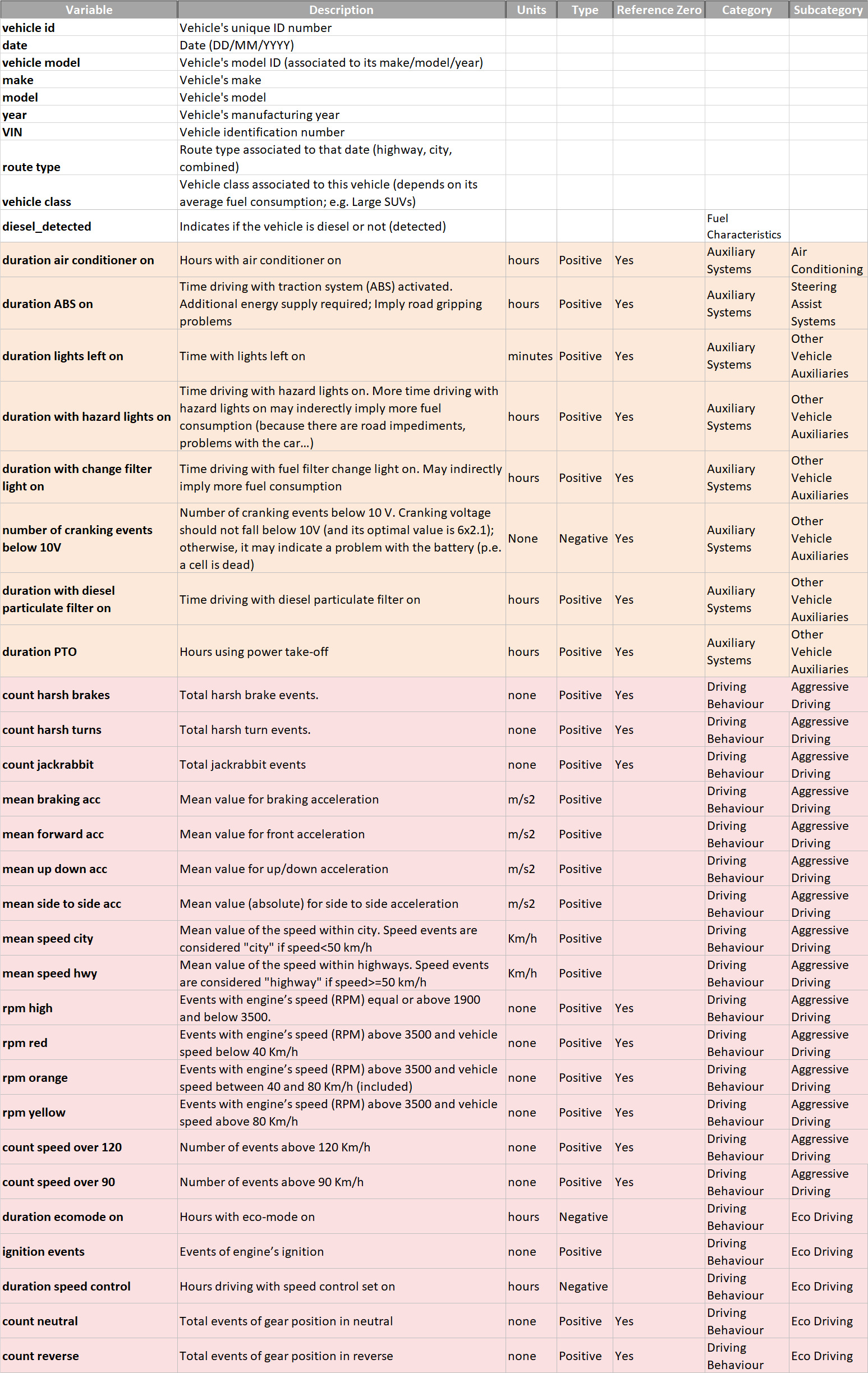}
  \end{tabular} 
  \caption{General variables and features used for predicting the fuel usage, with their associated categories and subcategories, according to \cite{zacharof2016review} for Auxiliary Systems and Driving Behaviour \label{table:FARUsedpt1}.}
\end{table}

\begin{table}[h!]
\centering
 \begin{tabular}{c@{\qquad}c@{\qquad}c}
\includegraphics[width=395pt, height=580pt, keepaspectratio]{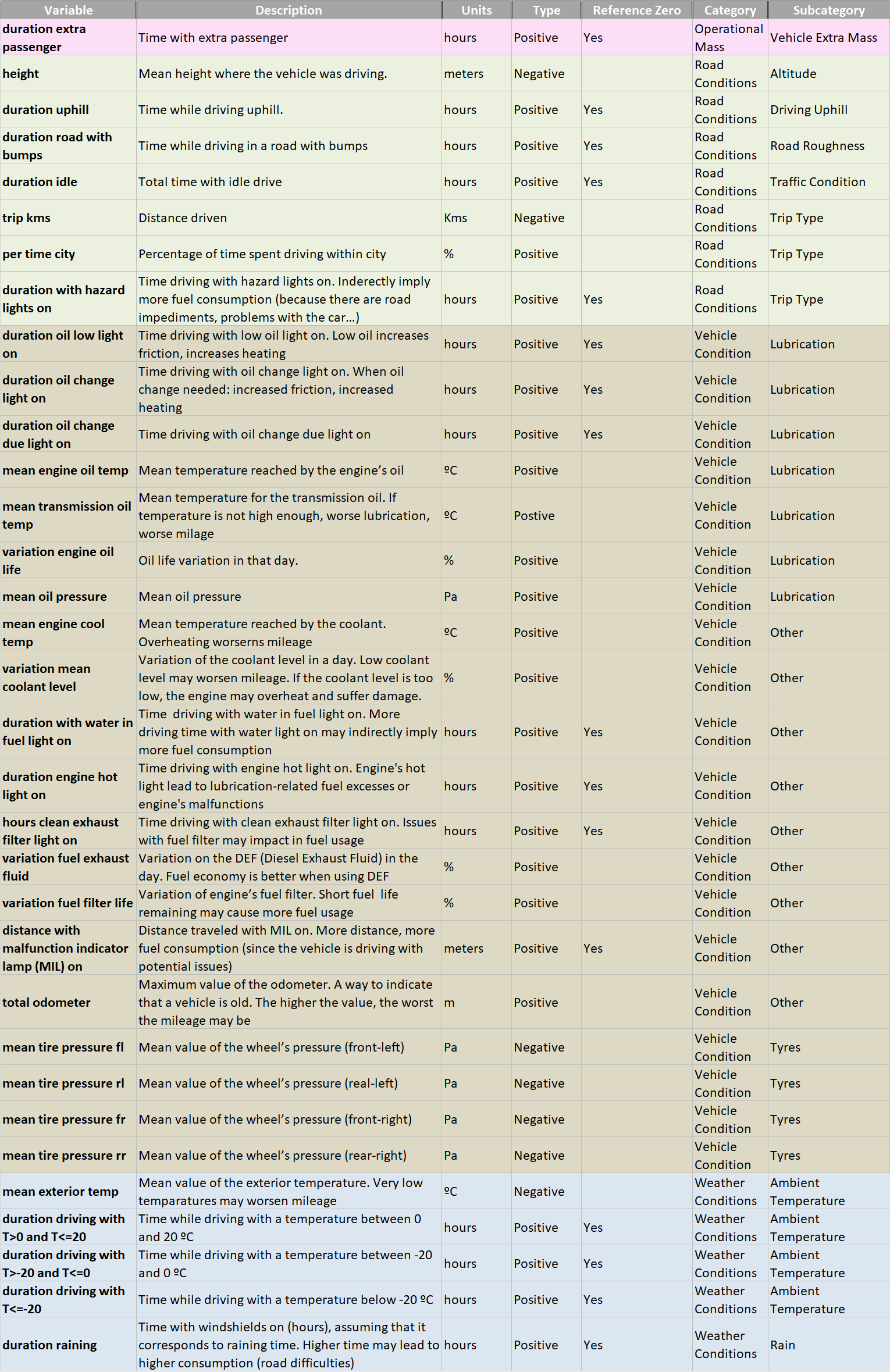}
  \end{tabular} 
  \caption{Features used for predicting the fuel usage, with their associated categories and subcategories, according to \cite{zacharof2016review} for Operational Mass, Road Conditions, Vehicle Conditions and Weather Conditions  \label{table:FARUsedpt2}}
\end{table}

\end{document}